\documentclass[twoside,11pt]{article}

\usepackage{jmlr2e}
\usepackage{amsmath}
\usepackage{algorithm}
\usepackage{algorithmic}
\usepackage{subfigure}
\usepackage{graphicx}
\usepackage{bbm}
\usepackage{tikz}
\usetikzlibrary{arrows}
\usetikzlibrary{graphs}
\usepackage{url}

\newcommand{\mb}{\mathbf}

\newcommand{\bs}{\boldsymbol}
\newcommand{\ie}{\textit{i.e.}}
\newcommand{\eg}{\textit{e.g.}}

\ShortHeadings{IFM LAB TUTORIAL SERIES \# 8, COPYRIGHT \copyright IFM LAB}{JIAWEI ZHANG, IFM LAB DIRECTOR}
\firstpageno{1}

\begin{document}

\title{Robot Basics: Representation, Rotation and Velocity}

\author{\name Jiawei Zhang \email jiawei@ifmlab.org \\
	\addr{Founder and Director}\\
       {Information Fusion and Mining Laboratory}\\
       (First Version: October 2022; Revision: November 2022.)}

\maketitle

%------------------------------------------------------------------------
\begin{abstract}

In this article, we plan to provide an introduction about some basics about robots for readers. Several key topics of classic robotics will be introduced, including \textit{robot representation}, \textit{robot rotational motion}, \textit{coordinates transformation} and \textit{velocity transformation}. By now, classic rigid-body robot analysis is still the main-stream approach in robot controlling and motion planning. In this article, no data-driven or machine learning based methods will be introduced. Most of the materials covered in this article are based on the rigid-body kinematics that the readers probably have learned from the physics course at high-school or college. Meanwhile, these classic robot kinematics analyses will serve as the foundation for the latest intelligent robot control algorithms in modern robotics studies.
\end{abstract}

%\textit{trajectory generation}, \textit{motion planning}, \textit{robot control}, \textit{robot manipulation} and \textit{diverse robot applications}

\begin{keywords}
Robotics; Robot Representation; Robot Configuration; Coordinate System; Rotational Motion; Linear Velocity; Angular Velocity
%; Trajectory Generation, Motion Planning; Robot Controlling; Robot Manipulation; Robot Applications\\
\end{keywords}

\tableofcontents

%------------------------------------------------------------------------
\section{Introduction}\label{}

According to the forecast report from UN (United Nations) \cite{un_ageing}, the world population aging will become one of the most challenging global problem of the 21st century. The working age population growth of the major developing countries (e.g., China, India) will gradually slow down and even start to decrease together with the major developed countries (e.g., UK, France, Germany, Japan and USA). To fulfill the ``tremendous gap'' between the supply and demand of working labor forces, various robots and automated machines has been (and will continue to be) developed and employed for massive production work globally. This is an irreversible trend for the 21st century. To accomplish such an objective, educating and training researchers and practitioner on robotics is imperative and critical at present.

\subsection{This Article}

In this tutorial article, we will provide a brief introduction about robotics for readers, which covers \textit{robot representation}, \textit{robot motion}, \textit{coordinate transformation} and \textit{velocity transformation} in the rotational motions. This tutorial article will also be organized according to the sequential order these topics. Some more advanced topics about \textit{robot kinematics}, \textit{trajectory generation}, \textit{robot control}, \textit{motion planning} and \textit{manipulation} and \textit{applications} will be introduced in the follow-up tutorial articles instead. Besides these tutorial articles, for readers interested in robot control, we also have several textbooks recommended for you to read as well, like \cite{10.5555/3100040, 10.5555/3165183}.

This tutorial article will be math-heavy. Just like you, I don't like math either, but we have to use math to model the real world. Based on the equations we introduce in the article, we will be able to write the code to represent and control the robot. The bitter has to been swallowed first, then the sweet will come afterwards. Before we discuss about the aforementioned topics, we would like to briefly describe the notations to be used in this article as follows.  

\subsection{Basic Notations}\label{subsec:basic_notation}

In the sequel of this article, we will use the lower case letters (e.g., $x$) to represent scalars, lower case bold letters (e.g., $\mb{x}$) to denote column vectors, bold-face upper case letters (e.g., $\mb{X}$) to denote matrices. Given a vector $\mb{x}$, its length is denoted as $\left\| \mb{x} \right\|$. Given a matrix $\mb{X}$, we denote $\mb{X}(i,:)$ and $\mb{X}(:,j)$ as its $i_{th}$ row and $j_{th}$ column, respectively. The ($i_{th}$, $j_{th}$) entry of matrix $\mb{X}$ can be denoted as either $\mb{X}(i,j)$ or $\mb{X}_{i,j}$, which will be used interchangeably. We use $\mb{X}^\top$ and $\mb{x}^\top$ to represent the transpose of matrix $\mb{X}$ and vector $\mb{x}$. The cross product of vectors $\mb{x}$ and $\mb{y}$ is represented as $\mb{x} \times \mb{y}$. A coordinate system is denoted as $\Sigma$, and the vector $\mb{x}$ in coordinate system $\Sigma$ can also be specified as $\mb{x}^{[\Sigma]}$. For a scalar $x$, vector $\mb{x}$ and matrix $\mb{X}$, we can also represent their first-order derivatives as $\dot{x}$, $\dot{\mb{x}}$ and $\dot{\mb{X}}$, and second-order derivatives as $\ddot{x}$, $\ddot{\mb{x}}$ and $\ddot{\mb{X}}$.

%------------------------------------------------------------------------
\section{Robot Representation}\label{sec:robot_representation}

In this tutorial article, we will focus on rigid-body robots with known shapes. Besides rigid-body robots, research on soft-body robots is also very popular nowadays, but it is out of the scope of this tutorial article.

%------------------------------------------------------------------------
%------------------------------------------------------------------------

\subsection{Robot Structure}

%------------------------
\begin{figure}[t]
    \centering
    \includegraphics[width=0.9\textwidth]{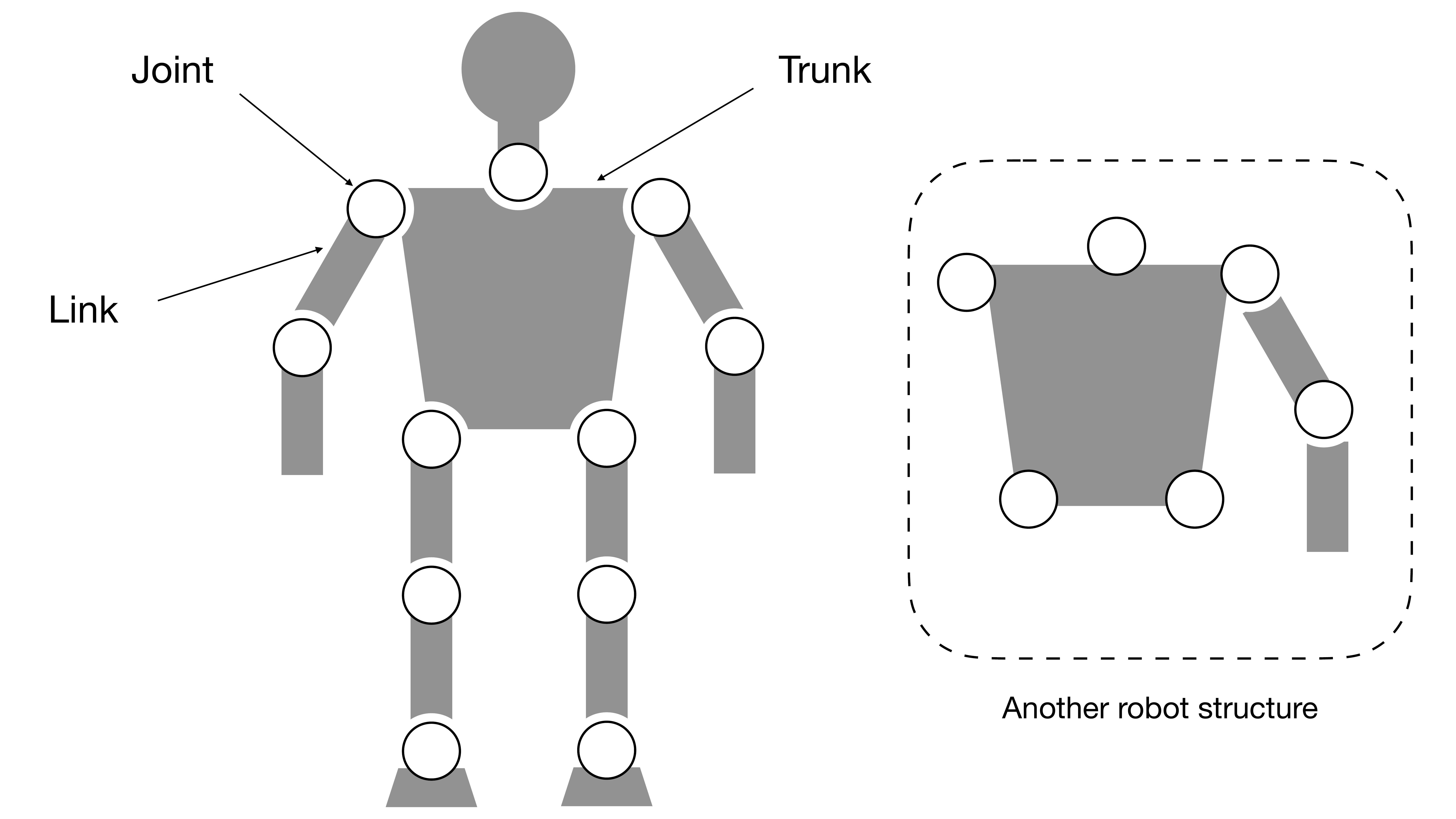}
    \caption{A Example of Rigid-Body Robot Structure (involving the trunk, joints and links).}
    \label{fig:robot_structure}
\end{figure}
%------------------------

%------------------------
\begin{figure}[t]
    \centering
    \includegraphics[width=0.9\textwidth]{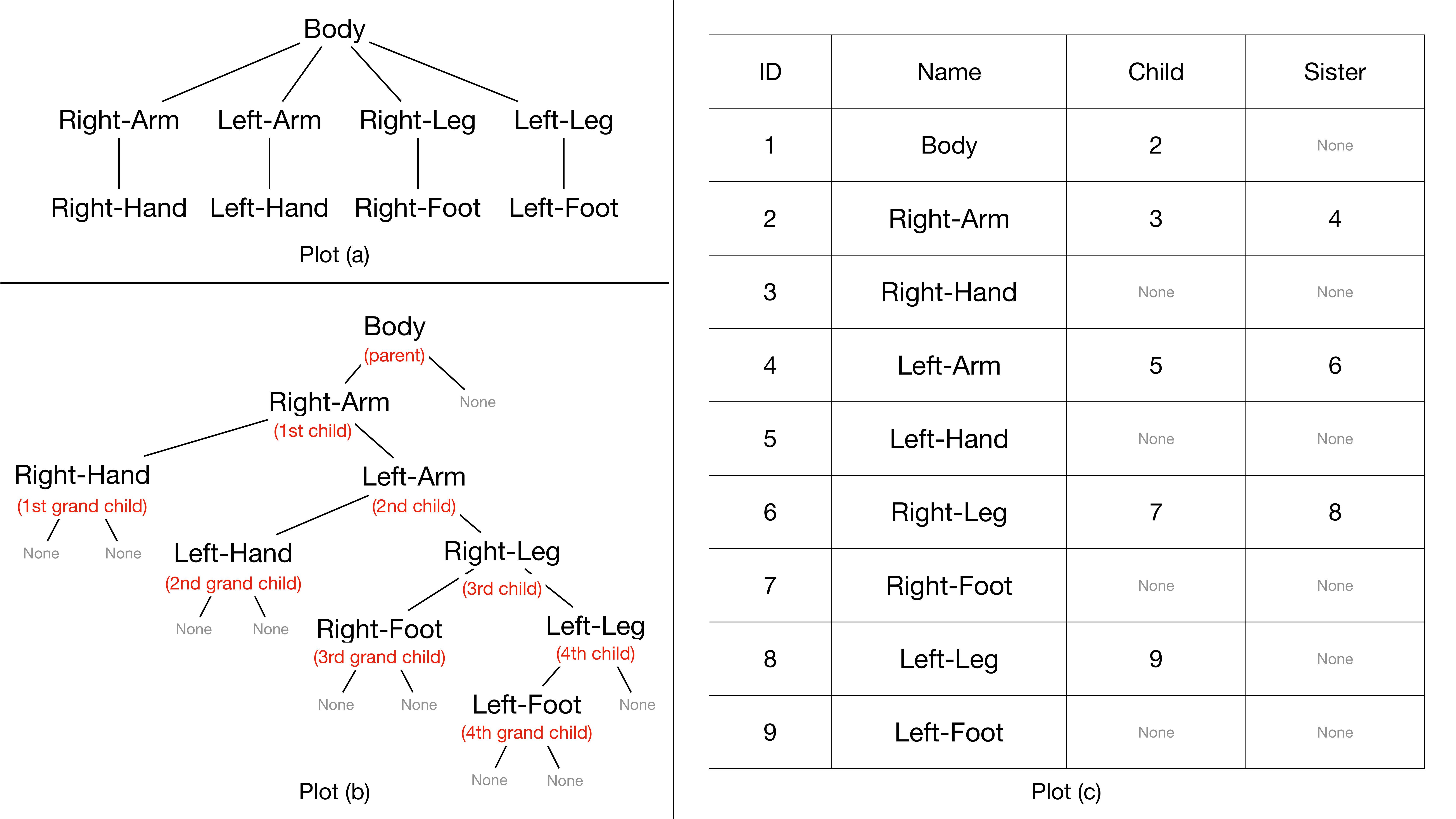}
    \caption{Data Structure for Representing Robot Body Structure: (a) balanced tree data structure, (b) binary tree data structure, (c) list of body parts.}
    \label{fig:robot_structure_data_structure}
\end{figure}
%------------------------

At the beginning, we would like to introduce the robot physical structure and its representation with certain data structures in program for readers.

%------------------------------------------------------------------------
\subsubsection{Robot Physical Structure}

As shown in Figure~\ref{fig:robot_structure}, we provide an example of the rigid-body robot structure, involving the trunk and a number of links connected together via the joints. For the representation and motion control simplicity, we prefer to attach the joints to the link that is further away from the trunk. In this robot structure, each link will have only one joint to control its motion, so the same algorithm can be applied to control all the links and joints in the robot. 

Meanwhile, in the small plot shown at the right-hand side in Figure~\ref{fig:robot_structure}, we also provide another robot structure, where the joints are attached to the links closer to the trunk. In such a structure, the robot trunk will have multiple attached motors, e.g., 2 from leg, 2 from arm and 1 from neck. For such a robot structure, the control of the robot trunk with $5$ attached joints will be very different from the control of the arms and legs, which may create extra difficulties in the robot motion and control.

%------------------------------------------------------------------------
\subsubsection{Data Structure}

To represent the robot structure shown in Figure~\ref{fig:robot_structure} in program, various data structures can be used to represent the robot body parts and connections among them. In robot control, we are especially interested in the parts that are moveable. As illustrated in the Plot (a) of Figure~\ref{fig:robot_structure_data_structure}, we can use a \textit{tree} to represent the robot structure, involving the body and four limbs directly connected to the trunk body. Meanwhile, each of the four limbs is also connected to either a hand node or a foot node as their child, respectively. 

For the nodes in Plot (a) of Figure~\ref{fig:robot_structure}, we observe that the tree nodes' out-degrees are quite different, {\eg}, the body root node's out degree is $4$, while the arm/leg nodes' out degrees are all $1$ instead. Such node out-degree inconsistency may create potential problems in writing the code for robot control.

In Plot (b) of Figure~\ref{fig:robot_structure}, we illustrate another tree structured representation of the robot physical structure, where each internal node is connected to two other nodes, {\ie}, one child and one sibling. For instance, the ``Right-Arm'' node is connected to both ``Right-Hand'' (its child node) and ``Left-Arm'' (its sibling node). For the nodes without children or siblings, we can just use the dummy ``None'' to fill the entries.

Compared with the balanced tree in Plot (a), the new tree structure in Plot (b) is skewed and inclined to the right-hand side. Meanwhile, the tree structure is a binary tree actually, as each node connects to two nodes at the lower level, programming based on which will be much easier. Actually, such a tree data structure can be implemented very easily as a list, where each node is represented as a record with unique IDs in the table, whose child and sibling nodes can be indicated as its attributes. An example to implement the binary tree structure as a list is also illustrated in Plot (c) of Figure~\ref{fig:robot_structure}.

%------------------------------------------------------------------------
\subsubsection{Notation Table}

Besides the basic notations introduced at the end of Section~\ref{subsec:basic_notation}, in this part, we will also introduce some other notations that will be used in the following sections for describing the robot parts and their motions in the following Table~\ref{tab:robot_link_notation}.

%------------------------
\begin{table}[H]
\centering
\small
\caption{Robot Link Notation in Math and Program}\label{tab:robot_link_notation}
\begin{tabular}{ |p{8cm}||p{3cm}|p{3cm}|  }
 \hline
 \multicolumn{3}{|c|}{Notation Table in Math and Program Code} \\
 \hline
Parameters of Robot Link& Math Notation & Code Notation \\
 \hline
 Link Self-ID & $j$ & self\_id \\
 Child ID	& N/A	& child\_id\\
Sibling ID & N/A & sibling\_id\\
Parent ID	& $i$ & parent\_id \\
Position Vector & $\mb{p}$ & $p$ \\
Rotation Matrix & $\mb{R}$ & $R$\\
Homogeneous Transition Matrix & $\mb{T}$ & $T$\\
Velocity in World Coordinate & $\mb{v}$ & v\\
Angular Velocity in World Coordinate & $\bs{\omega}$ & w\\
Joint Angle & $q$ & q\\
Joint Rotation Velocity & $\dot{q}$ & dq\\
Joint Rotation Acceleration & $\ddot{q}$ & ddq\\
Mass & $m$ & m\\
Center of Mass & ${\mb{c}}$ & c\\
Moment of Inertia &${\mb{I}}$ & I\\
 \hline
\end{tabular}
\end{table}
%------------------------

%------------------------------------------------------------------------
%------------------------------------------------------------------------
\subsection{Robot Configuration and Configuration Space}

In this part, we will introduce the concepts of \textit{robot configuration} and \textit{configuration space}. Both \textit{robot configuration} and \textit{configuration space} play an important role in robot motion, control and kinematics. In the following sections, when we introduce the \textit{robot motion}, \textit{coordinate transformation} and \textit{velocity transformation}, readers will observe that robot motion will involve a sequence of \textit{robot configurations} (more like a configuration path) connecting from the starting configuration to the desired target configuration within the pre-defined robot \textit{configuration space}.

%------------------------------------------------------------------------
\subsubsection{Robot Configuration}

For a rigid-body robot, its location and state can be uniquely specified by its \textit{configuration}. 

%------------------------
\begin{definition}
(\textbf{Robot Configuration}): The \textit{robot configuration} denotes a complete specification of the positions of every point in the robot system.
\end{definition}
%------------------------

In the real-world, we cannot enumerate all the points of a robot body to represent its \textit{configuration}. For rigid-body robots, since all its composition parts are rigid and have pre-known shapes, actually only a few numbers will be sufficient to represent its \textit{configuration}. 

%------------------------
\begin{figure}[t]
    \centering
    \includegraphics[width=0.9\textwidth]{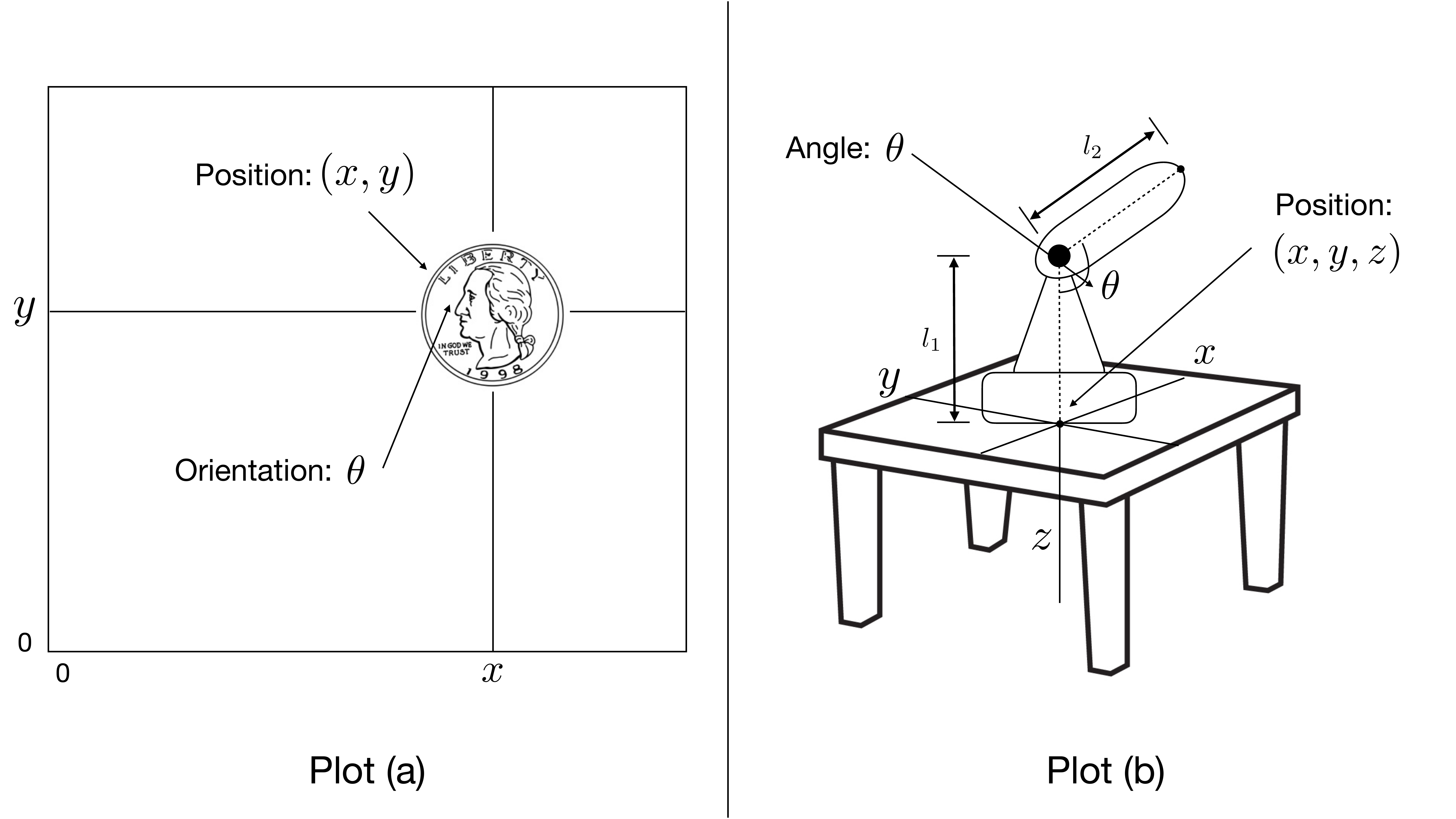}
    \caption{A Example of Rigid-Body Item Configuration.}
    \label{fig:configuration_example}
\end{figure}
%------------------------

%------------------------
\begin{example}
For instance, as shown in Plot (a) of Figure~\ref{fig:configuration_example}, given a coin lying heads up on a table (which can be represented as a 2D plane), the location and state of the coin can be specified with three parameters $\left( (x, y), \theta \right)$: (1) the coordinate parameter pair $(x, y)$ that denotes its location on the table, and (2) a third parameter $\theta$ that indicates the coin's orientation (head up or tail up).
\end{example}
%------------------------

%------------------------
\begin{example}
Besides the configuration in 2D space, 3D space is much more frequently used in robotics. As shown in Plot (b) of Figure~\ref{fig:configuration_example}, for a robot arm with one vertical link (of known length $l_1$) and a movable arm link (of known length $l_2$) connected by a joint, which forms an angle $\theta$ between these two links. The robot arm position and state can be represented with four parameters $\left( (x, y, z), \theta \right)$, where $(x, y, z)$ denotes the position of the base center and $\theta$ indicates the angle between these two arm links.
\end{example}
%------------------------

%------------------------------------------------------------------------
\subsubsection{Robot Configuration Space}

%------------------------
\begin{definition}
(\textbf{Configuration Space}): For a given robot, there usually exist different feasible configurations, and the set of all potential configurations of the robot defines the \textit{configuration space} of the robot.
\end{definition}
%------------------------

In some textbooks, the robot \textit{configuration space} is also called the \textit{C-space} for abbreviation. It is also easy to observe that given the robot's \textit{configuration space}, one of its specific \textit{configuration} will actually be a point in the \textit{configuration space}.

%------------------------
\begin{example}
For instance, for the coin shown in Plot (a) of Figure~\ref{fig:configuration_example}, let $\mathbbm{X}$ and $\mathbbm{Y}$ denote the sets of potential x-coordinate and y-coordinate values of the coin's position. Then, the \textit{configuration space} of the coin can be represented as $\mathbbm{X} \times \mathbbm{Y} \times \{Head, Tail\}$.
\end{example}
%------------------------

%------------------------
\begin{example}
When it comes to the robot arm shown in Plot (b) of Figure~\ref{fig:configuration_example}, let $\mathbbm{X}$, $\mathbbm{Y}$ and $\mathbbm{Z}$ denote the sets of potential $x$, $y$ and $z$ coordinate values of the robot arm base center, the \textit{configuration space} of the robot arm can be represented as $\mathbbm{X} \times \mathbbm{Y} \times \mathbbm{Z} \times [0, 2\pi)$, where the arms' angle $\theta$ can be any value within the range of $[0, 2\pi)$.
\end{example}
%------------------------

In the real world, there exist different ways to define robot configuration space, and we use the sphere as an example to illustrate different configuration spaces below.

\begin{example}
For instance, a point on a sphere surface of radius $r \in \mathbbm{R}$ can be uniquely represented with its ``(latitude, longitude)'' pair, just like the GPS localization on the earth. Meanwhile, the point on the sphere surface can also be represented with a ($x$, $y$, $z$) coordinate pair in the Euclidean space subject to the constraint $x^2+y^2+z^2=r^2$. 
\end{example}

The former representation in the above example is normally called the \textit{explicit representation}, while the latter one is called the \textit{implicit representation}. Based on the \textit{robot configuration} and \textit{configuration space} concepts introduced above, next we will study robot representation and coordinate rotational transformation based on the \textit{implicit representation} in Euclidean space by default.

%------------------------------------------------------------------------
%------------------------------------------------------------------------

\subsection{Robot Coordinate System and Representation}

Based on the above descriptions, we know that robot's position and state can be uniquely represented with its \textit{configuration}, which denotes a group of numerical parameters. In this part, we will introduce the robot coordinate system to clearly specify such parameters. The coordinate systems to be used for representing robots include both the \textit{world coordinates} and \textit{local coordinates}.

%------------------------------------------------------------------------

\subsubsection{World Coordinate System}\label{subsubsec:world_coordinate_system}

To control robots in the real world, we need a fixed coordinate system to define the positions of all parts of the robot, {\ie}, the \textit{world coordinate}.

%------------------------
\begin{figure}[t]
    \centering
    \includegraphics[width=0.9\textwidth]{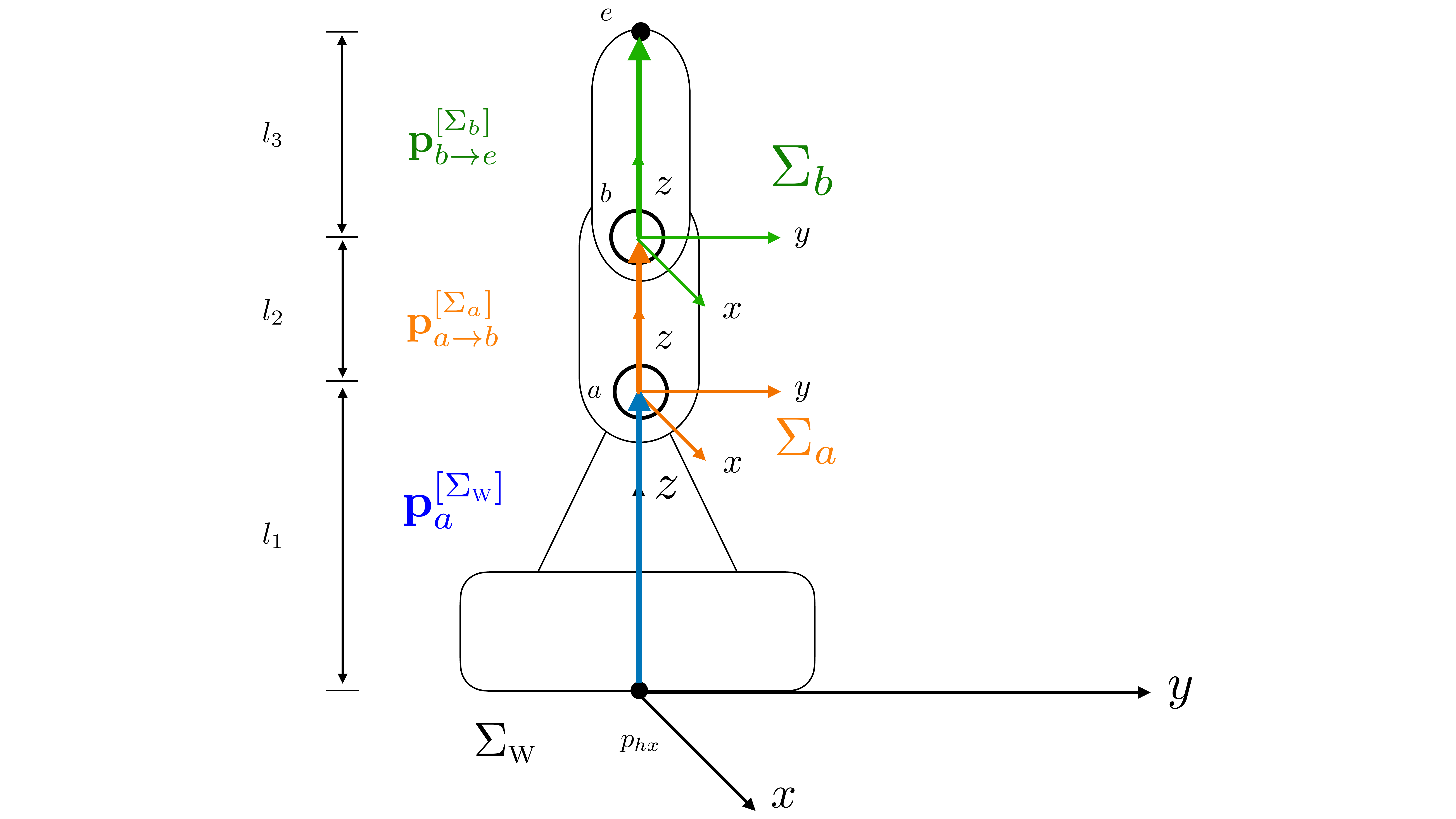}
    \caption{A Example of World Coordinate.}
    \label{fig:world_configuration_example}
\end{figure}
%------------------------

\begin{definition}
(\textbf{World Coordinate}): Formally, the \textit{world coordinate} denotes a coordinate system with a fixed origin and fixed orientations of the $x$, $y$ and $z$ axes. From the robot perspective, the orientation of the $x$ axis faces forward, $y$ axis to the left and $z$ axis faces up. In this tutorial article, we can refer to the \textit{world coordinate} with a default notation $\Sigma_{\textsc{w}}$.
\end{definition}

For the \textit{world coordinate} notation $\Sigma_{\textsc{w}}$, the term $\Sigma$ denotes the coordinate and the subscript $\textsc{w}$ differentiates the \textit{world coordinate} from other \textit{local coordinates} to be introduced later.

\begin{example}
As shown in Figure~\ref{fig:world_configuration_example}, we provide an example of the world coordinate $\Sigma_{\textsc{w}}$, whose origin is located at the center of the robot arm base. The robot arm has two links connected via the joints $a$ and $b$, and the end point of the second arm link is denoted as $e$.

Based on the world coordinate $\Sigma_{\textsc{w}}$, we can represent the vector point from the origin to the first joint $a$ as a vector $\mb{p}_a^{[\Sigma_{\textsc{w}}]}$ (in the blue color), whose orientation faces upward. Similarly, we can also represent the vectors pointing from the world coordinate system origin to the joint $b$ and end point $e$ as $\mb{p}_b^{[\Sigma_{\textsc{w}}]}$ and $\mb{p}_e^{[\Sigma_{\textsc{w}}]}$, respectively.
\end{example}

In the above example, we observe the length of the arm links are $l_1$ (from the base to the first joint), $l_2$ (from the first joint to the second one) and $l_3$ (from the second joint to the end point). Therefore, the vectors $\mb{p}_{o \to a}^{[\Sigma_{\textsc{w}}]}$, $\mb{p}_{o \to b}^{[\Sigma_{\textsc{w}}]}$ and $\mb{p}_{o \to e}^{[\Sigma_{\textsc{w}}]}$ can actually be represented as 
\begin{equation}
\mb{p}_{o \to a}^{[\Sigma_{\textsc{w}}]} = (0,0,l_1)^\top \text{ ,\ \ } \mb{p}_{o \to b}^{[\Sigma_{\textsc{w}}]} = (0,0,l_1+l_2)^\top \text{ ,\ \ } \mb{p}_{o \to e}^{[\Sigma_{\textsc{w}}]} = (0,0,l_1+l_2+l_3)^\top,
\end{equation}
where the subscript $o \to a$ denotes the vector pointing from the origin of the world coordinate to joint $a$. 

When there exists one single world coordinate in controlling the robot, we can just omit the $o$ in representing vectors pointing from the origin to certain points and vector representation $\mb{p}_{o \to a}^{[\Sigma_{\textsc{w}}]}$ can be simplified as $\mb{p}_{a}^{[\Sigma_{\textsc{w}}]}$ in the remaining parts of this tutorial. Similarly, vectors $\mb{p}_{o \to b}^{[\Sigma_{\textsc{w}}]}$ and $\mb{p}_{o \to e}^{[\Sigma_{\textsc{w}}]}$ can also be simplified as $\mb{p}_{b}^{[\Sigma_{\textsc{w}}]}$ and $\mb{p}_{e}^{[\Sigma_{\textsc{w}}]}$.

%------------------------------------------------------------------------
\subsubsection{Local Coordinate Systems}

Besides the \textit{world coordinate}, we can also introduce various \textit{local coordinate systems} attached to different parts of the robot rigid body. For the servo or motor driven robots, we normally prefer attach the origins of the \textit{local coordinate systems} to the rotor centers the servo/motor.

\begin{definition}
(\textbf{Local Coordinate}): By picking one point of the robot rigid body as the origin and specifying the $x$, $y$ and $z$ axes, we can construct a \textit{local coordinate system} for a robot part. The \textit{local coordinate} defines the position and state of the robot body parts from a local perspective.
\end{definition}

In the \textit{local coordinate systems}, the representation of robot position and state will be very different from what it looks like from the world coordinate

\begin{example}
As shown in Figure~\ref{fig:world_configuration_example}, by taking the robot arm joint $a$ as the origin, we can also build a \textit{local coordinate system} $\Sigma_a$ (in the orange color). The orientation of the $x$, $y$ and $z$ axes are identical to the axis orientations of the \textit{world coordinate} $\Sigma_{\textsc{w}}$. In a similar way, we can introduce the \textit{local coordinate system} $\Sigma_b$ (in the green color) with the identical axis orientations and have the origin attached to joint $b$.

Within the \textit{local coordinate system} $\Sigma_a$, we can represent the robot arm pointing from joint $a$ to the second joint $b$ as a vector 
\begin{equation}
\mb{p}_{a \to b}^{[\Sigma_a]} = (0, 0, l_2)^\top,
\end{equation} 
where $l_2$ denotes the length of the corresponding arm link. 

Similarly, in the \textit{local coordinate} $\Sigma_b$, we can represent the robot arm pointing from joint $b$ to the end point $e$ as a vector 
\begin{equation}
\mb{p}_{b \to e}^{[\Sigma_b]} = (0, 0, l_3)^\top,
\end{equation} 
where $l_3$ denotes the arm link length.
\end{example}

From the numbers of the vector representations of $\mb{p}_a^{[\Sigma_{\textsc{w}}]}$, $\mb{p}_b^{[\Sigma_{\textsc{w}}]}$ and $\mb{p}_e^{[\Sigma_{\textsc{w}}]}$ in the world coordinate system, as well as $\mb{p}_b^{[\Sigma_a]}$ and $\mb{p}_e^{[\Sigma_b]}$ in the local coordinate systems, it seems that
\begin{align}
\mb{p}_b^{[\Sigma_{\textsc{w}}]} &= \mb{p}_a^{[\Sigma_{\textsc{w}}]} + \mb{p}_{a \to b}^{[\Sigma_a]},\\
\mb{p}_e^{[\Sigma_{\textsc{w}}]} &= \mb{p}_b^{[\Sigma_{\textsc{w}}]} + \mb{p}_{b \to e}^{[\Sigma_b]} = \mb{p}_a^{[\Sigma_{\textsc{w}}]} + \mb{p}_{a \to b}^{[\Sigma_a]} + \mb{p}_{b \to e}^{[\Sigma_b]}.
\end{align}
Meanwhile, the above equations can hold as the $x$, $y$ and $z$ axes orientations of the local coordinate systems $\Sigma_a$ and $\Sigma_b$ are identical to those of the world coordinate $\Sigma_{\textsc{w}}$, i.e., $\mb{p}_{a \to b}^{[\Sigma_a]} = \mb{p}_{a \to b}^{[\Sigma_{\textsc{w}}]}$ and $\mb{p}_{b \to e}^{[\Sigma_b]} = \mb{p}_{b \to e}^{[\Sigma_{\textsc{w}}]}$. Meanwhile, when the arm links rotates, the local coordinate orientations will change and necessary transformation needs to be added to the above equations.

%------------------------------------------------------------------------
%------------------------------------------------------------------------
%------------------------------------------------------------------------

\section{Robot End Point Position}

Based on the above robot representation and notations, in this section, we will discuss about the robot end point position before and after the rotation of the robot arm. To interact with the world, robot end points play an important role. In this section, we are especially interested in the position, while the velocity of the end point will discussed in the following Section~\ref{sec:robot_velocity_torque} instead.

\subsection{Single-Link Robot End Point Position}

In operation, robot links will rotate around the joint, and will change the position and state of the robot in the space. Such rotational movement can be clearly described with the \textit{rotation matrix} and \textit{homogeneous transformation matrix}. In this part, we will first introduce the \textit{rotation matrix} that can denote the rotation transformation. After that, we will introduce the \textit{homogeneous transformation matrix} that can project any vector from one coordinate system to another.

%------------------------------------------------------------------------
\subsubsection{Rotation Matrix}\label{subsubsec:rotation_matrix}

%------------------------
\begin{figure}[t]
    \centering
    \includegraphics[width=0.9\textwidth]{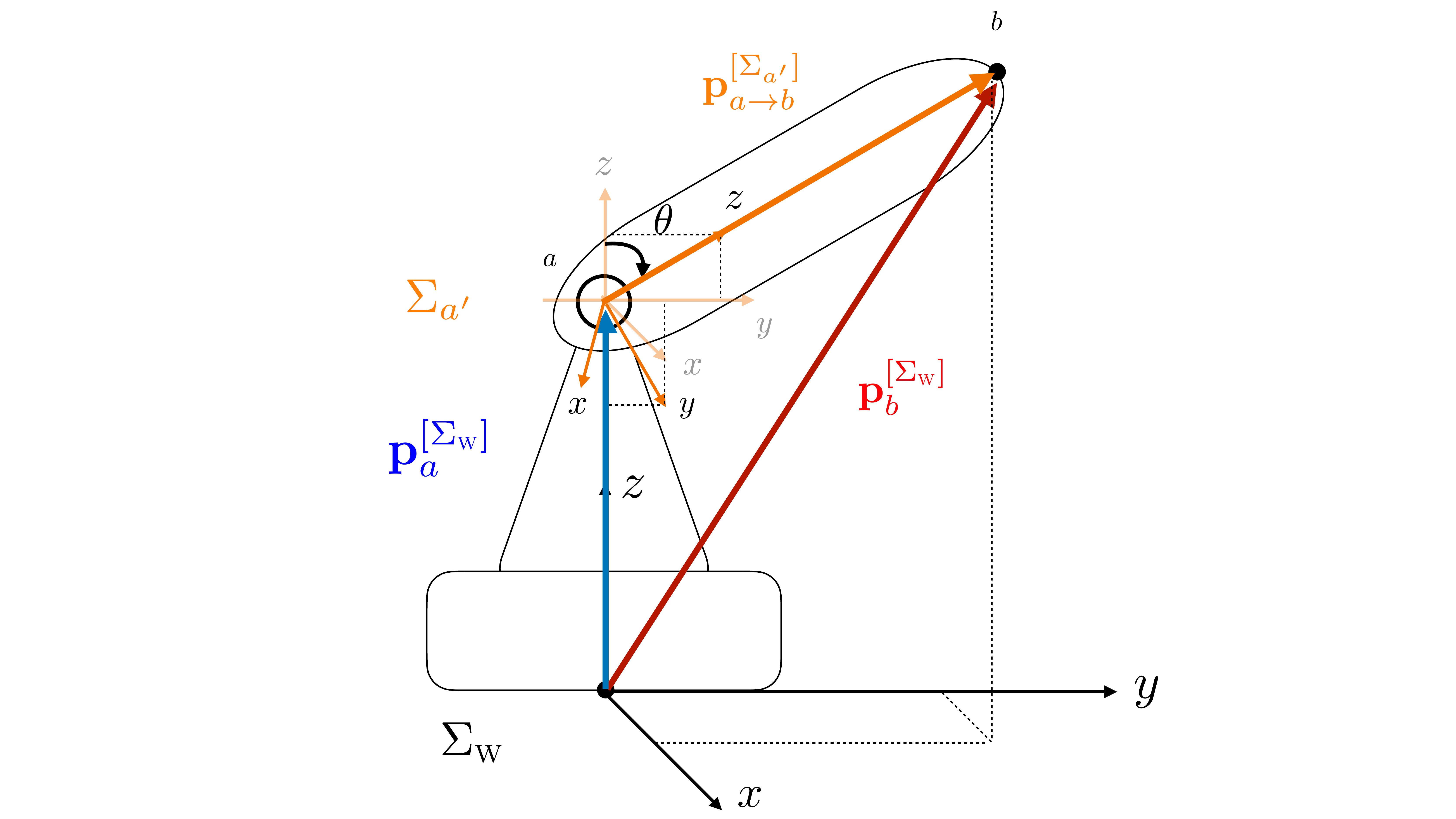}
    \caption{An Example of Robot Arm Local Coordinate After Rotation.}
    \label{fig:local_configuration_example}
\end{figure}
%------------------------

Let's first discard the second robot arm link from the example shown in Figure~\ref{fig:world_configuration_example} and only use the 1-arm robot illustrated in Figure~\ref{fig:local_configuration_example} as an example.

%------------------------
\begin{example}
As shown in Figure~\ref{fig:local_configuration_example}, we rotate the robot arm around joint $a$ from the previous vertical position in Figure~\ref{fig:world_configuration_example} in the clockwise direction with an angle of $\theta$ degrees to the current position. As the arm link rotates, the \textit{local coordinate} attached to the robot arm joint will rotate clockwise with $\theta$ degrees as well, and the new local coordinate system after the rotation can be denoted as $\Sigma_{a'}$.

Within the \textit{local coordinate system} $\Sigma_{a'}$, the representations of robot arm link pointing from joint $a$ to $b$ after the rotation are not changed actually. In other words, in the new local coordinate $\Sigma_{a'}$, the arm link vector after rotation can still be represented as 
\begin{equation}
\mb{p}_{a \to b}^{[\Sigma_{a'}]} = (0, 0, l_2)^\top.
\end{equation}

Meanwhile, looking from the \textit{world coordinate} $\Sigma_{\textsc{w}}$, we observe that the arm link vector pointing from joint $a$ to $b$ will be different from $\mb{p}_{a \to b}^{[\Sigma_{a'}]}$ actually, which can be represented as 
\begin{equation}
\mb{p}_{a \to b}^{[\Sigma_{\textsc{w}}]} = (0, l_2 \sin \theta, l_2 \cos \theta)^\top.
\end{equation}
\end{example}
%------------------------

To further clearly illustrate the rotations of the local coordinate systems from $\Sigma_a$ to $\Sigma_{a'}$, we first need to talk about the relationship between the local coordinate systems $\Sigma_a$ and $\Sigma_{a'}$ with the world coordinate $\Sigma_{\textsc{w}}$. Since the initial local coordinate system $\Sigma_a$ axis orientations are identical to those of $\Sigma_{\textsc{w}}$, their conversion into each other is super easy and can be done just by shifting the origins to a new position without any rotation transformation. Next, we will focus on the rotation transformation from $\Sigma_{a'}$ to $\Sigma_{\textsc{w}}$, which will be identical to the rotation transformation from $\Sigma_{a'}$ to $\Sigma_{a}$.

By projecting the $x$, $y$ and $z$ axes of $\Sigma_{a'}$ to $\Sigma_{\textsc{w}}$, the unit vectors parallel to the $x$, $y$ and $z$ axes of $\Sigma_{a'}$ within the world coordinate $\Sigma_{\textsc{w}}$ can be represented as
\begin{equation}
\mb{e}^{[\Sigma_{a'} \to \Sigma_{\textsc{w}}]}_{x} = (1, 0, 0)^\top \text{ , } \mb{e}^{[\Sigma_{a'} \to \Sigma_{\textsc{w}}]}_{y} = (0, \cos \theta, -\sin \theta)^\top \text{ , } \mb{e}^{[\Sigma_{a'} \to \Sigma_{\textsc{w}}]}_{z} = (0, \sin \theta, \cos \theta)^\top. 
\end{equation}

%------------------------
\begin{definition}
(\textbf{Rotation Matrix}): Based on the above unit vectors projected from $\Sigma_{a'}$ to $\Sigma_{\textsc{w}}$, we can define the ration matrix from $\Sigma_{a'}$ to $\Sigma_{\textsc{w}}$ as:
\begin{equation}\label{equ:rotation_matrix}
\mb{R}^{[\Sigma_{a'} \to \Sigma_{\textsc{w}}]} = (\mb{e}^{[\Sigma_{a'} \to \Sigma_{\textsc{w}}]}_{x}, \mb{e}^{[\Sigma_{a'} \to \Sigma_{\textsc{w}}]}_{y}, \mb{e}^{[\Sigma_{a'} \to \Sigma_{\textsc{w}}]}_{z}) = \begin{bmatrix}
1 & 0 & 0 \\
0 & \cos \theta & \sin \theta \\
0 & -\sin \theta & \cos \theta \\
\end{bmatrix}.
\end{equation}
\end{definition}
%------------------------

With the \textit{rotation matrix} $\mb{R}^{[\Sigma_a \to \Sigma_{\textsc{w}}]}$, we can actually project the arm link vector pointing from the joint $a$ to joint $b$ from the local coordinate $\Sigma_{a'}$ to its representation in the world coordinate $\Sigma_{\textsc{w}}$:
\begin{equation}
\underbrace{\begin{bmatrix}
0\\
l_2 \sin \theta\\
l_2 \cos \theta\\
\end{bmatrix}}_{\mb{p}_{a \to b}^{[\Sigma_{\textsc{w}}]}} = \underbrace{\begin{bmatrix}
1 & 0 & 0 \\
0 & \cos \theta & \sin \theta \\
0 & -\sin \theta & \cos \theta \\
\end{bmatrix}}_{\mb{R}^{[\Sigma_{a'} \to \Sigma_{\textsc{w}}]} }
\underbrace{\begin{bmatrix}
0\\
0\\
l_2\\
\end{bmatrix}}_{\mb{p}_{a \to b}^{[\Sigma_{a'}]}}.
\end{equation}

%------------------------------------------------------------------------

\subsubsection{Homogeneous Transformation Matrix}

Next, let's have a look at the vector pointing from the world coordinate system origin to the robot arm joint $b$ after the rotation. Looking from the world coordinate, we can represent the vector as
\begin{align}
\mb{p}_b^{[\Sigma_{\textsc{w}}]} &= \mb{p}_a^{[\Sigma_{\textsc{w}}]} + \mb{p}_{a \to b}^{[\Sigma_{\textsc{w}}]} \\
&= \mb{p}_a^{[\Sigma_{\textsc{w}}]} + \mb{R}^{[\Sigma_{a'} \to \Sigma_{\textsc{w}}]} \mb{p}_{a \to b}^{[\Sigma_{a'}]}.
\end{align}

Linear algebra will be frequently used in representing robots in this article, and the above formula can also be rewritten as follows instead:
\begin{equation}\label{equ:end_point_position_representation}
\begin{bmatrix}
\mb{p}_b^{[\Sigma_{\textsc{w}}]}\\
1
\end{bmatrix} = 
\begin{bmatrix}
\mb{R}^{[\Sigma_{a'} \to \Sigma_{\textsc{w}}]} & \mb{p}_a^{[\Sigma_{\textsc{w}}]}\\
\mb{0} & 1
\end{bmatrix}
\begin{bmatrix}
\mb{p}_{a \to b}^{[\Sigma_{a'}]}\\
1
\end{bmatrix}
\end{equation}
Some extra entries with dummy values ($0$ or $1$) are added to the vectors and matrices in the above formula just to make their dimensions match with each other. 

\begin{definition}
(\textbf{Homogeneous Transformation Matrix}): In the above equation, we introduce a new matrix representation composed of the position vector $\mb{p}_a$ and the rotation matrix $\mb{R}_{a'}$, which is also named as the \textit{homogeneous transformation matrix}, {\ie},
\begin{equation}
\mb{T}^{[\Sigma_{a'} \to \Sigma_{\textsc{w}}]} = \begin{bmatrix}
\mb{R}^{[\Sigma_{a'} \to \Sigma_{\textsc{w}}]} & \mb{p}_a^{[\Sigma_{\textsc{w}}]}\\
\mb{0} & 1
\end{bmatrix}
\end{equation}
\end{definition}

The robot \textit{homogeneous transformation matrix} $\mb{T}^{[\Sigma_{a'} \to \Sigma_{\textsc{w}}]}$ will convert all points described in the robot arm's local coordinate to the world coordinate. Therefore, we can generally say that the \textit{homogeneous transformation matrix} $\mb{T}^{[\Sigma_{a'} \to \Sigma_{\textsc{w}}]}$ describes the position and state of the robot arm. Since the robot rotation transformation is very common, to simplify the notations, we can just use $\Sigma_a$ to represent all the coordinate systems with origin at joint $a$ (including $\Sigma_a$ and $\Sigma_{a'}$ discussed before). As to $\Sigma_a$ is the coordinate system before or after the rotation, readers can differentiate them from the context by yourself.

%------------------------------------------------------------------------
%------------------------------------------------------------------------
\subsection{Multi-Link Robot End Point Position}\label{subsec:multi_link_robot_transformation}

Real-world robots are usually composed of multiple links connected to each other by the joints. At the end of the first section, we will take a more general robot with two links to study the local coordinate transformation via the chain rule.

%------------------------------------------------------------------------
\subsubsection{Multi-Link Robots with Local to Local Coordinate Systems}

\begin{figure}[t]
    \centering
    \includegraphics[width=0.9\textwidth]{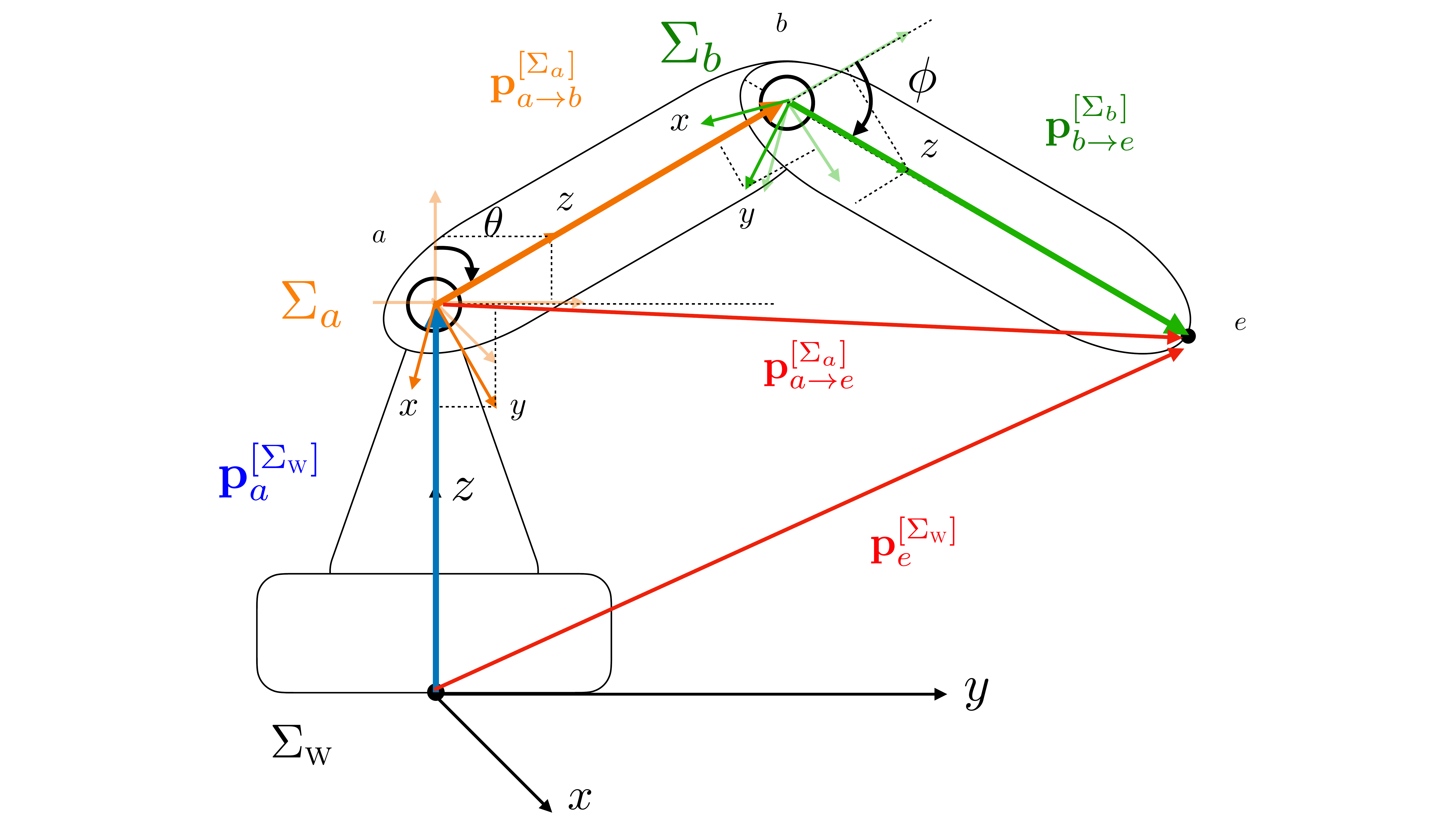}
    \caption{A Example of Multi-Link Robot Arm Coordinate Transformation via Chain Rule.}
    \label{fig:local_configuration_example_chain_rule}
\end{figure}

As illustrated in Figure~\ref{fig:local_configuration_example_chain_rule}, we rotates both of the two arm links of the robot shown in Figure~\ref{fig:world_configuration_example} with certain degrees: the first robot arm rotates clockwise with an angle of $\theta$ degrees around joint $a$, while the second link rotates clockwise with an angle $\phi$ degrees around joint $b$. At these two joints, we introduce two local coordinate systems $\Sigma_a$ and $\Sigma_b$, and these two arm links are defined as vectors $\mb{p}_{a \to b}^{[\Sigma_a]}$ and $\mb{p}_{b \to e}^{[\Sigma_b]}$ within these two local coordinate systems, respectively. 

According to the analysis in the previous parts, by taking $\Sigma_a$ as the ``world coordinate'', we can represent the rotation matrix to project coordinate $\Sigma_b$ to $\Sigma_a$ as
\begin{equation}
\mb{R}^{[\Sigma_b \to \Sigma_a]} = (\mb{e}_x^{[\Sigma_b\to\Sigma_a]}, \mb{e}_y^{[\Sigma_b\to\Sigma_a]}, \mb{e}_z^{[\Sigma_b\to\Sigma_a]}),
\end{equation}
where the unit vectors are defined as follows:
\begin{equation}
\mb{e}^{[\Sigma_b\to\Sigma_a]}_{x} = (1, 0, 0)^\top \text{; } \mb{e}^{[\Sigma_b\to\Sigma_a]}_{y} = (0, \cos \phi, - \sin \phi)^\top \text{; } \mb{e}^{[\Sigma_b\to\Sigma_a]}_{z} = (0, \sin \phi, \cos \phi)^\top. 
\end{equation}

%------------------------
%------------------------

Meanwhile, for the end point of the second arm link, we can represent its position in the coordinate $\Sigma_a$ as $\mb{p}_{a \to e}^{[\Sigma_a]}$, whose representation after the rotation can be represented as
\begin{equation}\label{equ:local_end_point_position_representation}
\begin{bmatrix}
\mb{p}_{a \to e}^{[\Sigma_a]}\\
1
\end{bmatrix} = 
\underbrace{ \begin{bmatrix}
\mb{R}^{[\Sigma_b\to\Sigma_a]} & \mb{p}_{a \to b}^{[\Sigma_a]}\\
\mb{0} & 1
\end{bmatrix}}_{\mb{T}^{[\Sigma_b \to \Sigma_a]}}
\begin{bmatrix}
\mb{p}_{b \to e}^{[\Sigma_b]}\\
1
\end{bmatrix}.
\end{equation}

The matrix that accomplishes the transformation denotes the \textit{homogeneous transformation matrix} $\mb{T}^{[\Sigma_b\to\Sigma_a]}$ from coordinate $\Sigma_b$ to $\Sigma_a$.

%------------------------------------------------------------------------
\subsubsection{Chain Rule on Homogeneous Transformation}

By combining Equation~\ref{equ:end_point_position_representation} and Equation~\ref{equ:local_end_point_position_representation}, we can further derive the end point's position of the second link in the world coordinate to be
\begin{equation}\label{equ:chain_rule_end_point_position_representation}
\begin{bmatrix}
\mb{p}_e^{[\Sigma_{\textsc{w}}]}\\
1
\end{bmatrix} = \mb{T}^{[\Sigma_a\to\Sigma_{\textsc{w}}]} \begin{bmatrix}
\mb{p}_{a \to e}^{[\Sigma_a]}\\
1
\end{bmatrix} = \mb{T}^{[\Sigma_a\to\Sigma_{\textsc{w}}]} \mb{T}^{[\Sigma_b\to\Sigma_a]} \begin{bmatrix}
\mb{p}_{b \to e}^{[\Sigma_b]}\\
1
\end{bmatrix}.
\end{equation}

%-----
On the right-hand side of the equation, we multiply the two \textit{homogeneous transformation matrices} $\mb{T}^{[\Sigma_a\to\Sigma_{\textsc{w}}]}$ and $\mb{T}^{[\Sigma_b\to\Sigma_a]}$ together to transform the local vector representation $\mb{p}_{b \to e}^{[\Sigma_b]}$ in coordinate $\Sigma_b$ to the world coordinate $\mb{p}_e^{[\Sigma_{\textsc{w}}]}$. Formally, we can also represent the multiplication of $\mb{T}^{[\Sigma_a\to\Sigma_{\textsc{w}}]} \mb{T}^{[\Sigma_b\to\Sigma_a]}$ as the \textit{homogeneous transformation matrix} from the local coordinate $\Sigma_{b}$ to the world coordinate $\Sigma_{\textsc{w}}$:
\begin{equation}
\mb{T}^{[\Sigma_b \to \Sigma_{\textsc{w}}]} = \mb{T}^{[\Sigma_a\to\Sigma_{\textsc{w}}]} \mb{T}^{[\Sigma_b\to\Sigma_a]},
\end{equation}
and the Equation~\ref{equ:chain_rule_end_point_position_representation} can also be rewritten as
\begin{equation}
\begin{bmatrix}
\mb{p}_e^{[\Sigma_\textsc{w}]}\\
1
\end{bmatrix} = \mb{T}^{[\Sigma_b \to \Sigma_{\textsc{w}}]} 
\begin{bmatrix}
\mb{p}_{b \to e}^{[\Sigma_b]}\\
1
\end{bmatrix}.
\end{equation}

For the robot arms with more than two links, the end point's position can be derived in a similar way with the chain rule on the multiple homogeneous transformation matrices between coordinate systems. By now, we should have provide a detailed introduction on multi-link rigid-body robot end point position and transformation across coordinate systems for the readers already. In the next section, we will discuss more about the robot rotational motion and its relationship with the robot velocity for the readers.

%------------------------------------------------------------------------
\section{Robot Rotation}

In the previous section, we have introduced the robot coordinate systems and representations. We have also briefly talk about the robot arm rotation, as well as the robot position and state changes due to the rotation movement. In this section, we will provide a more detailed analysis about the robot's general rotational motion and illustrate several other important properties of the rotation matrix defined before.

%------------------------------------------------------------------------
%------------------------------------------------------------------------

\subsection{General Rotation}

The rotation movements we introduce in the previous section are around one certain axis (e.g., the $x$ axis) and the rotation direction is clockwise. In this part, we will introduce a more general robot rotation around any axes with either clockwise and counter-clockwise directions.

%------------------------------------------------------------------------

\subsubsection{Rotation Axis}

%------------------------
\begin{figure}[t]
    \centering
    \includegraphics[width=0.9\textwidth]{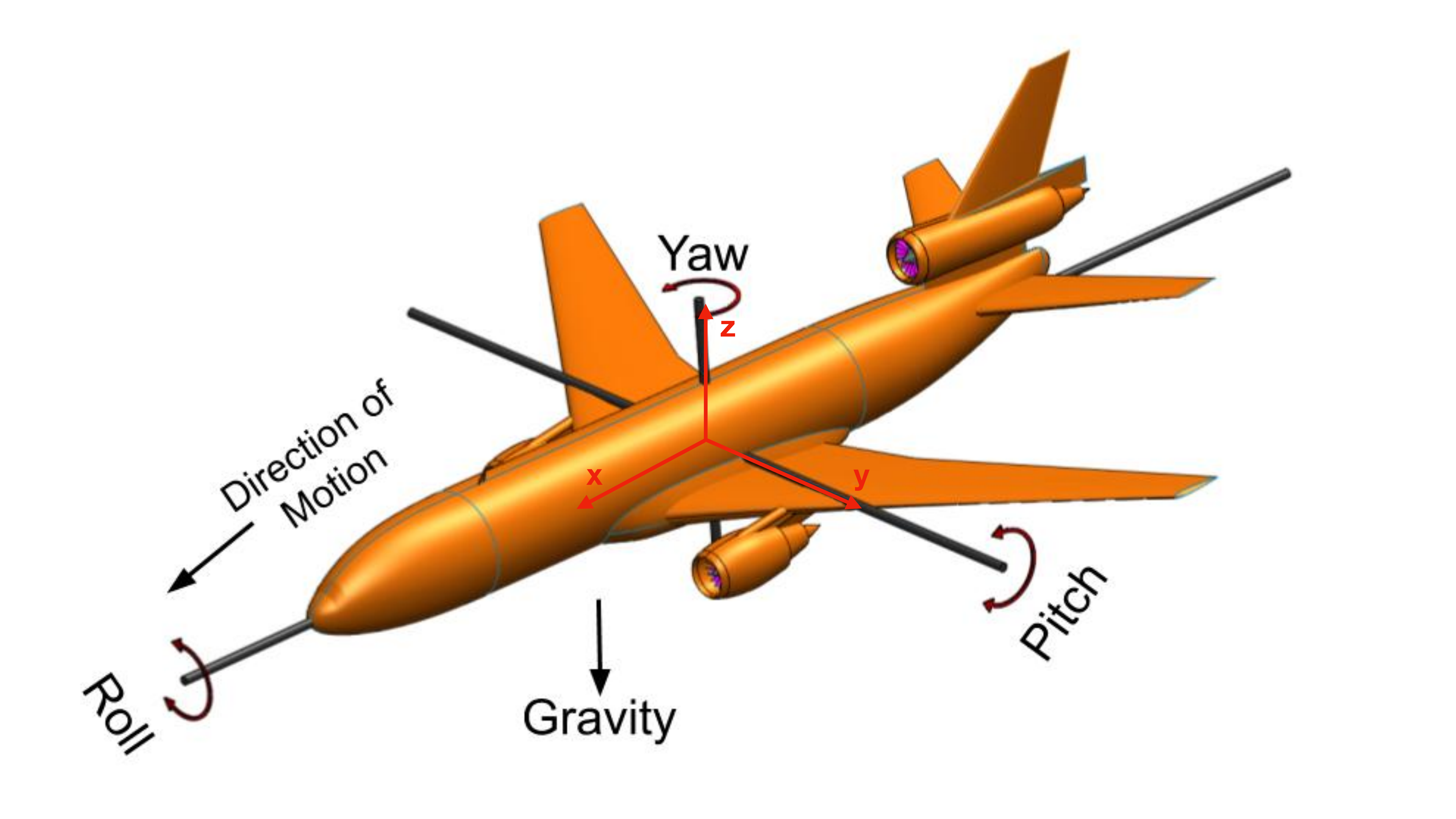}
    \caption{A Example of Roll, Pitch and Yaw Rotation Movements.}
    \label{fig:roll_pitch_yaw}
\end{figure}
%------------------------

Depending on the orientation of the joints, robots can rotate around the joints not only along the $x$, $y$ and $z$ axes but also a combination of them, which will lead to the rotation in any directions in the 3D space. 

\begin{definition}
(\textbf{Roll, Pitch and Yaw}): Specifically, for the rotations around the $x$, $y$ and $z$ axes, we can give them different names to describe the rotation:
\begin{itemize}
\item \textit{Roll}: The rotation movement along the $x$ axis is also called \textit{roll}.
\item \textit{Pitch}: The rotation movement along the $y$ axis is also called \textit{pitch}.
\item \textit{Yaw}: The rotation movement along the $z$ axis is also called \textit{yaw}.
\end{itemize}

In the \textit{roll} rotational motion, for all the points in the robot body, their $x$ coordinate is not changed, while their $y$ and $z$ coordinates will change together with the rotation. Meanwhile, for the \textit{pitch} rotation, all the $x$ and $z$ coordinates will change due to the rotation. The \textit{yaw} rotation will change the $x$ and $y$ coordinates instead.
\end{definition}

Readers may also wonder why the rotation along these axes are named as \textit{roll}, \textit{pitch} and \textit{yaw}, respectively. These movement names are actually borrowed from the airplane control. As illustrated in Figure~\ref{fig:roll_pitch_yaw}, given a airplane, based on the coordinate system definition provided in Section~\ref{subsubsec:world_coordinate_system}, we can introduce a coordinate system along the plane body ($x$ axis faces forward, $y$ axis to the airplane left and $z$ axis faces up).

If the readers having experiences in driving airplane, the two \textit{ailerons} on the outer rear edge of each wing cause the plane to roll to the left or right. The \textit{elevator} on the horizontal tail surface controls the pitch movement of the plane. Meanwhile, the rudder on the vertical tail fin controls swivels from side to side, which will push the tail in a left or right direction and control the \textit{yaw} movement of the plane. Similar names of the motions are also used to describe the robot rotation here.

%------------------------------------------------------------------------

\subsubsection{Rotation Matrices for Roll, Pitch and Yaw}

For the \textit{roll}, \textit{pitch} and \textit{yaw} rotational motions, to differentiate them from each other, we can denote their rotation angles as $\phi$, $\theta$ and $\psi$, respectively. Here, we assume the rotations are all in the counter-clockwise direction (i.e., the positive angles). Based on the degrees and the representations we learn from Section~\ref{subsubsec:rotation_matrix}, regardless of the coordinate systems, we can represent the corresponding rotation matrices for \textit{roll}, \textit{pitch} and \textit{yaw} as follows:
\begin{align}\label{equ:rpy_rotation_matrices}
\mb{R}_x(\phi) &= \begin{bmatrix}
1 & 0 & 0\\
0 & \cos \phi & - \sin \phi\\
0 & \sin \phi & \cos \phi\\
\end{bmatrix}, \\
\mb{R}_y(\theta) &= \begin{bmatrix}
\cos \theta & 0 & \sin \theta\\
0 & 1 & 0\\
- \sin \theta & 0 & \cos \theta\\
\end{bmatrix}, \\
\mb{R}_z(\psi) &= \begin{bmatrix}
\cos \psi & -\sin \theta & 0\\
\sin \psi & \cos \theta & 0\\
0 & 0 & 1\\
\end{bmatrix}. \\
\end{align}

Let's assume, given a vector $\mb{p}$ pointing from the current joint to the end point, if we rotate the current joint with a $\phi$-degree angle around the $x$ axis, a $\theta$-degree angle around the $y$ axis and a $\psi$-degree angle around the $z$ axis, we can represent the vector after those 3-way rotations as
\begin{equation}
\mb{p}' = \mb{R}_z(\psi) \mb{R}_y(\theta) \mb{R}_x(\phi) \mb{p} = \mb{R}_{rpy}(\phi, \theta, \psi) \mb{p},
\end{equation}
where the \textit{roll-pitch-yaw} rotation matrix $\mb{R}_{rpy}(\phi, \theta, \psi)$ can be represented as follows:
\begin{align}\label{equ:rpy_rotation_matrix}
&\mb{R}_{rpy}(\phi, \theta, \psi) = \\
&\begin{bmatrix}
\cos \psi \cos \theta & - \sin \psi \cos \phi + \cos \psi \sin \theta \sin \phi & \sin \psi \sin \phi + \cos \psi \sin \theta \cos \phi \\
\sin \psi \cos \theta & \cos \psi \cos \phi + \sin \psi \sin \theta \sin \phi & - \cos \psi \sin \phi + \sin \psi \sin \theta \cos \phi \\
- \sin \theta & \cos \theta \sin \phi & \cos \theta \cos \phi\\
\end{bmatrix}.
\end{align}

The readers may have notice the rotation matrix $\mb{R}_x(\phi)$ defined in Equation~\ref{equ:rpy_rotation_matrices} is slightly different from the rotation matrix defined in Equation~\ref{equ:rotation_matrix}. The main differences lie at the rotation direction, which actually follows the right-hand rule. With the right hand index finger, middle finger, ring finger and little finger following the rotation direction, if the thumb orientation is the same as the axis positive direction, then the rotation angle will be positive; otherwise, the rotation angle will be negative.

According to the right-hand rule, for the rotations in the counter-clockwise direction, their rotation matrices can be simply represented as those in Equation~\ref{equ:rpy_rotation_matrices}. Meanwhile, when it comes to the clockwise rotation (just like Figure~\ref{fig:local_configuration_example}), we can add a negative mark to the angles before applying them in Equation~\ref{equ:rpy_rotation_matrices} for defining the rotation matrices, which will lead to the rotation matrix representation in Equation~\ref{equ:rotation_matrix}.

With the \textit{roll-pitch-yaw} rotation matrix $\mb{R}_{rpy}(\phi, \theta, \psi)$ defined above, arbitrary rotation can be achieved by the rotation angle pair $(\phi, \theta, \psi)$ around the $x$, $y$ and $z$ axes, respectively.

%------------------------------------------------------------------------
%------------------------------------------------------------------------

\subsection{Rotation Matrix Property}\label{subsec:rotation_matrix_property}

\vspace{10pt}
\begin{theorem}
Given a rotation matrix $\mb{R}$, it will be \textit{orthogonal}, {\ie},
\begin{equation}
\mb{R}\mb{R}^\top = \mb{I}.
\end{equation}
\end{theorem}
\vspace{10pt}

\begin{proof}
When deriving the \textit{rotation matrix} in Equation~\ref{equ:rotation_matrix}, we mention that the rotation matrix can be represented as a group of unit vectors corresponding to the $x$, $y$ and $z$ axes. For instance, the rotation matrix along the $x$ axis in the counter-clockwise direction with an angle of $\phi$ degrees can be represented as
\begin{equation}
\mb{R}_x(\phi) = \begin{bmatrix}
1 & 0 & 0\\
0 & \cos \phi & - \sin \phi\\
0 & \sin \phi & \cos \phi\\
\end{bmatrix} = (\mb{e}_x, \mb{e}_y, \mb{e}_z).
\end{equation}

We can observe that the product of $\mb{R}_x(\phi)^\top \mb{R}_x(\phi)$ will lead to an identity matrix, since
\begin{align}
\mb{R}_x(\phi)^\top \mb{R}_x(\phi) &= \begin{bmatrix}
\mb{e}_x^\top\\
\mb{e}_y^\top\\
\mb{e}_z^\top\\
\end{bmatrix} (\mb{e}_x, \mb{e}_y, \mb{e}_z)\\
&= \begin{bmatrix}
\mb{e}_x^\top\mb{e}_x & \mb{e}_x^\top\mb{e}_y & \mb{e}_x^\top\mb{e}_z\\
\mb{e}_y^\top\mb{e}_x & \mb{e}_y^\top\mb{e}_y & \mb{e}_y^\top\mb{e}_z\\
\mb{e}_z^\top\mb{e}_x & \mb{e}_z^\top\mb{e}_y & \mb{e}_z^\top\mb{e}_z\\
\end{bmatrix}\\
&= \begin{bmatrix}
1 & 0 & 0\\
0 & 1 & 0\\
0 & 0 & 1\\
\end{bmatrix}.
\end{align}

For the rotation matrices along the $y$ and $z$ axes, it is also easy to obtain that
\begin{equation}
\mb{R}_y(\theta)^\top \mb{R}_y(\theta) = \mb{I} \in \mathbbm{R}^{3 \times 3}, \mb{R}_z(\psi)^\top \mb{R}_z(\psi) = \mb{I} \in \mathbbm{R}^{3 \times 3}.
\end{equation}

How about the \textit{roll-pitch-yaw} rotation matrix $\mb{R}_{rpy}(\phi, \theta, \psi)$ introduced in Equation~\ref{equ:rpy_rotation_matrix}? Since $\mb{R}_{rpy}(\phi, \theta, \psi) = \mb{R}_z(\psi) \mb{R}_y(\theta) \mb{R}_x(\phi)$, we can get
\begin{align}
&\mb{R}_{rpy}(\phi, \theta, \psi)^\top \mb{R}_{rpy}(\phi, \theta, \psi) \\
& = \left( \mb{R}_z(\psi) \mb{R}_y(\theta) \mb{R}_x(\phi) \right)^\top \left( \mb{R}_z(\psi) \mb{R}_y(\theta) \mb{R}_x(\phi) \right)\\
& = \mb{R}_x(\phi)^\top \mb{R}_y(\theta)^\top \mb{R}_z(\psi)^\top \mb{R}_z(\psi) \mb{R}_y(\theta) \mb{R}_x(\phi) \\
& = \mb{R}_x(\phi)^\top  \left( \mb{R}_y(\theta)^\top \left( \mb{R}_z(\psi)^\top \mb{R}_z(\psi) \right) \mb{R}_y(\theta) \right) \mb{R}_x(\phi) \\
& = \mb{I}.
\end{align}

Meanwhile, readers can also just calculate the product of $\mb{R}_{rpy}(\phi, \theta, \psi)^\top \mb{R}_{rpy}(\phi, \theta, \psi)$ based on the concrete representations shown in Equation~\ref{equ:rpy_rotation_matrix} on a piece of paper, whose result should also be a identity matrix with value $1$ on the diagonal.
\end{proof}

Here, if we assume the rotation matrix $\mb{R}$ is invertible, by multiplying both sides of equation $\mb{R} \mb{R}^\top = \mb{I}$ with $\mb{R}^{-1}$, we can obtain
\begin{equation}
\mb{R}^\top = \mb{R}^{-1}.
\end{equation}

%------------------------------------------------------------------------
%------------------------------------------------------------------------

\subsection{Angular Velocity}\label{subsec:angular_velocity}

%------------------------
\begin{figure}[t]
    \centering
    \includegraphics[width=0.9\textwidth]{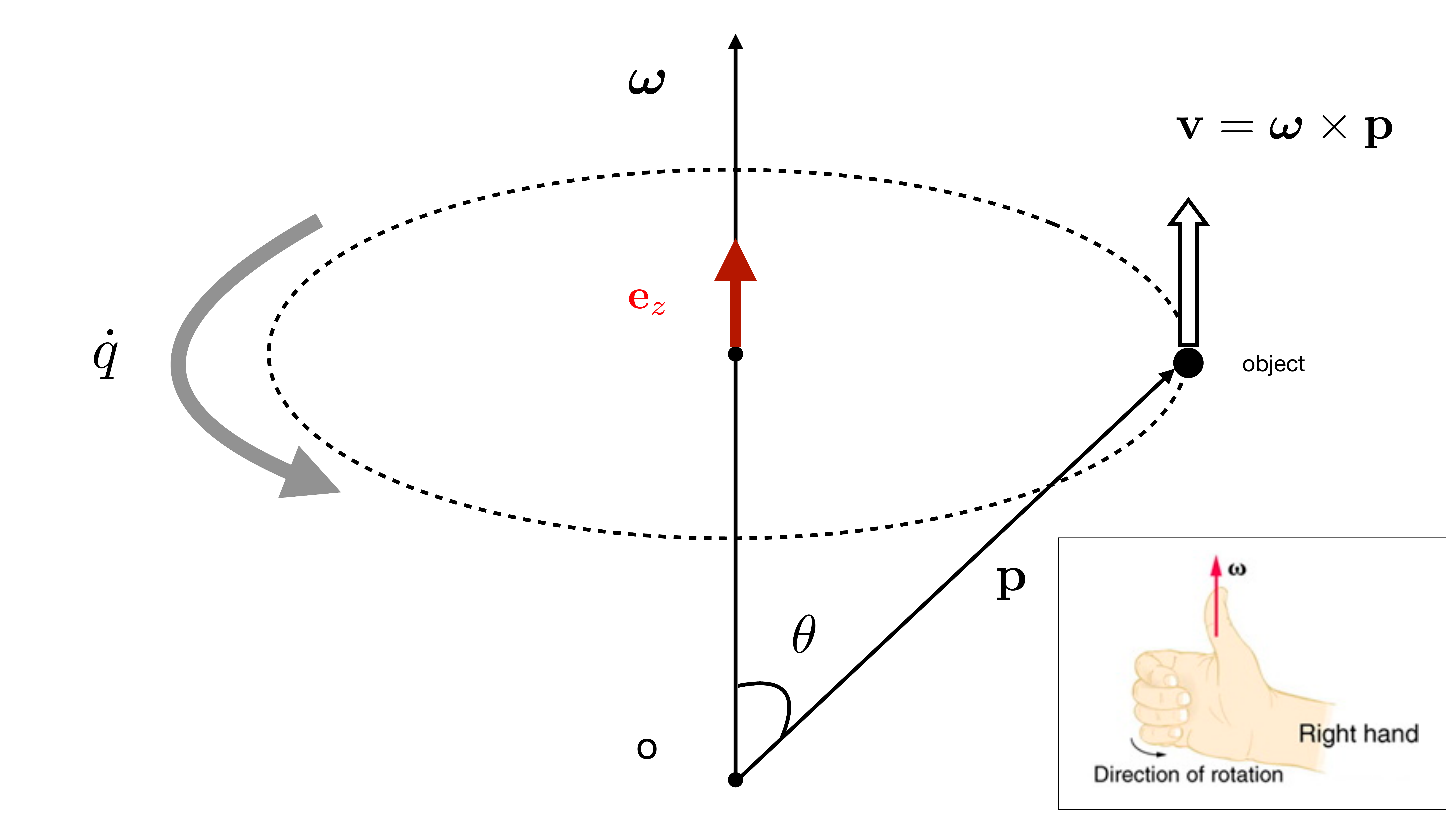}
    \caption{Angular Velocity Vector (Right Hand Rule).}
    \label{fig:angular_velocity}
\end{figure}
%------------------------

Before we talk more about the rotation matrix, we would like to have a cut-in subsection here to talk about the robot rotation \textit{angular velocity} first. More information about velocity and transformation will be provided in the following Section~\ref{sec:robot_velocity_torque}.

%------------------------------------------------------------------------

\subsubsection{Angular Velocity Vector}

As shown in Figure~\ref{fig:angular_velocity}, given an object rotating in the counter clockwise direction around the $z$ axis, according to the right-hand rule, we can represent the angular velocity vector orientation by the unit vector $\mb{e}_z$. Meanwhile, the rotation speed is denoted by a scalar $\dot{q} = 1$ rad/s. For the angular velocity unit ``rad/s'', it is equivalent to an ordinary frequency of $\frac{1}{2 \pi}$ hertz or cycles per second. So, if an object rotation angular velocity is $2 \pi$ rad/s, it will rotate 1 cycle per second.

\vspace{10pt}
\noindent \textbf{Some Remarks}: In this article, we like to use a regular scalar $q$ or vector $\mb{p}$ to represent the angular and position of an object, and use $\dot{q}$ and $\dot{\mb{p}}$ to represent its first-order derivatives ({\ie}, velocity), and $\ddot{q}$, $\ddot{\mb{p}}$ to represent the second-order derivatives ({\ie}, acceleration).
\vspace{10pt}

In the 3D space, similar to the velocity and positions, the object rotation angular velocity can also be represented as a vector. For this example, based on the angular velocity direction unit vector $\mb{e}_z$ and rotation velocity $\dot{q}$, we can represent the object rotation angular velocity can be represented as:
\begin{equation}\label{equ:angular_velocity}
\bs{\omega} = \dot{q} \mb{e}_z = \begin{bmatrix}
0\\
0\\
1\\
\end{bmatrix},
\end{equation}
where each element has the unit of rad/s.

%------------------------------------------------------------------------

\subsubsection{Linear Velocity Vector}

In Figure~\ref{fig:angular_velocity}, we also show the object position, which is denoted by a vector $\mb{p}$ from the origin pointing to its current position, and the angle between $\mb{p}$ and the rotation axis $\bs{\omega}$ is $\theta$. For the object in rotation, its \textit{linear velocity} can be represented as a vector denoted as the \textit{cross product} of its \textit{angular velocity vector} and its current \textit{position vector}. 

%------------------------
\begin{figure}[t]
    \centering
    \includegraphics[width=0.6\textwidth]{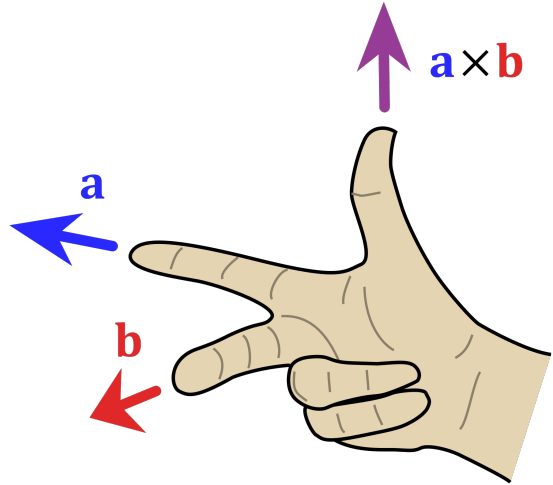}
    \caption{The Right-Hand Rule Indicating the Direction of the Cross Product Result Vector. (Given two vectors $\mb{a}$ and $\mb{b}$ represented by the index finger and middle finger, their cross-product result $\mb{a} \times \mb{b}$ direction is represented by the thumb finger, which is perpendicular to both $\mb{a}$ and $\mb{b}$.)}
    \label{fig:cross_product_right_hand_rule}
\end{figure}
%------------------------

For the object with angular velocity $\bs{\omega}$ and the position vector $\mb{p}$, we can represent its \textit{linear velocity vector} as
\begin{equation}
\mb{v} = \bs{\omega} \times \mb{p}.
\end{equation}

\vspace{5pt}
\noindent \textbf{Background Knowledge}: Given two vectors $\mb{a}$ and $\mb{b}$, their cross product result $\mb{a} \times \mb{b}$ will also be a vector that it perpendicular to both $\mb{a}$ and $\mb{b}$. To determine the direction of the result vector, as indicated in Figure~\ref{fig:cross_product_right_hand_rule}, it also follows the right hand rule. We just need to place the index finger to vector $\mb{a}$ and middle finger to vector $\mb{b}$, then the thumb finger will be the direction of $\mb{a} \times \mb{b}$. Following this rule, we can also observe that $\mb{a} \times \mb{b} = - \mb{b} \times \mb{a}$.
\vspace{5pt}

With the right hand rule introduced above, we can decide the direction of the \textit{linear velocity vector} $\mb{v}$, and observe that $\mb{v} \perp \bs{\omega}$ and $\mb{v} \perp \mb{p}$. As to the length of vector $\mb{v}$, it can be represented to be
\begin{equation}
\left\| \mb{v} \right\| = \left\| \bs{\omega} \right\| \left\| \mb{p} \right\| \sin \theta.
\end{equation}

If we know the elements of the vectors $\bs{\omega}$ and $\mb{p}$, we can also represent their cross product to be
\begin{equation}\label{equ:velocity_vector_representation}
\bs{\omega} \times \mb{p} = \begin{bmatrix}
\omega_x\\
\omega_y\\
\omega_z\\
\end{bmatrix} \times \begin{bmatrix}
p_x\\
p_y\\
p_z\\
\end{bmatrix} = \begin{bmatrix}
\omega_y p_z - \omega_z p_y\\
\omega_z p_x - \omega_x p_z\\
\omega_x p_y - \omega_y p_x\\
\end{bmatrix}.
\end{equation}

We observe that the result vector of $\bs{\omega} \times \mb{p}$ can actually also be obtained via a matrix-vector multiplication shown as follows:
\begin{equation}
\begin{bmatrix}
\omega_y p_z - \omega_z p_y\\
\omega_z p_x - \omega_x p_z\\
\omega_x p_y - \omega_y p_x\\
\end{bmatrix} = \begin{bmatrix}
0 & -\omega_z & \omega_y \\
\omega_z & 0 & -\omega_x \\
-\omega_y & \omega_x & 0 \\
\end{bmatrix} 
\begin{bmatrix}
p_x\\
p_y\\
p_z\\
\end{bmatrix}.
\end{equation}
In the following subsection, we will illustrate the physical meanings of the matrix multiplied with the vector $\mb{p}$, which can be build from the angular velocity vector directly.

%------------------------------------------------------------------------

\subsubsection{Rotation Transformation on Velocity Vectors}

If the origin in Figure~\ref{fig:angular_velocity} denotes a robot joint, which may rotate with time, the object's \textit{angular velocity vector} together with the central axis will also change with the rotation. Here, we can represent the rotation as matrix $\mb{R}$. By multiplying both sides of Equation~\ref{equ:angular_velocity} with matrix $\mb{R}$, we can get
\begin{equation}
\mb{R} \bs{\omega} = \mb{R}  \dot{q} \mb{e}_z.
\end{equation}

If we introduce two new vectors $\bs{\omega}'=\mb{R} \bs{\omega}$ and $\mb{e}_z'=\mb{R}\mb{e}_z$, then the above equation can be rewritten as follows:
\begin{equation}
\bs{\omega}' = \dot{q} \mb{e}_z',
\end{equation}
where $\bs{\omega}'$ denotes the new angular velocity vector and $\mb{e}_z'$ is the new axis. The velocity scalar value is still $\dot{q}$ rad/s/. Similarly, by introducing another new vector $\mb{v}'=\mb{R}\mb{v}$, we can also represent the object's linear velocity vector after the rotation as
\begin{equation}
\mb{v}' = \mb{R}\mb{v} = \mb{R} (\bs{\omega} \times \mb{p}) = (\mb{R} \bs{\omega}) \times (\mb{R} \mb{p}) = \bs{\omega}' \times \mb{e}_z'.
\end{equation}

These above velocity transformation equations and analysis results will be frequently used in calculating the robot end point velocity in the following Section~\ref{sec:robot_velocity_torque}.

%------------------------------------------------------------------------

\subsubsection{Rotation Matrix and Angular Velocity Vector}

Before we have introduced that via the rotation matrix, we can map a robot arm link from a local coordinate system, {\eg}, $\bar{\mb{p}}$, to the world coordinate system as follows:
\begin{equation}
\mb{p} = \mb{R} \bar{\mb{p}}.
\end{equation}

For robot arm, its link vector representation $\bar{\mb{p}}$ within the local coordinate system is static and has a fixed representation, which can be denoted as
\begin{equation}
\mb{R}^\top \mb{p} = \mb{R}^\top \mb{R} \bar{\mb{p}} \Rightarrow \bar{\mb{p}} = \mb{R}^\top \mb{p}.
\end{equation}

Meanwhile, both $\mb{p}$ and $\mb{R}$ will change with time. If we take the derivatives of $\mb{p}$ and $\mb{R}$ with regard to time, we can obtain
\begin{equation}
\dot{\mb{p}} = \dot{\mb{R}} \bar{\mb{p}} = \dot{\mb{R}} \left( \mb{R}^\top \mb{p} \right) =  \left( \dot{\mb{R}} \mb{R}^\top \right) \mb{p},
\end{equation}
where $\dot{\mb{p}} = \frac{\mathrm{d} \mb{p}}{\mathrm{d} t}$ and $\dot{\mb{R}} = \frac{\mathrm{d} \mb{R}}{\mathrm{d} t}$ represents the changing rate of the object's position and its rotation matrix.

Meanwhile, for an object, its position changing rate is actually its velocity, {\ie}, 
\begin{equation}
\dot{\mb{p}} = \mb{v} = \bs{\omega} \times \mb{p},
\end{equation}
which will lead to an important equation shown as follows
\begin{equation}\label{equ:velocity_rotation_matrix_relationship_equation}
\bs{\omega} \times \mb{p} = \left( \dot{\mb{R}} \mb{R}^\top \right) \mb{p}.
\end{equation}
The above equation correlates the object's \textit{angular velocity vector} $\bs{\omega}$ and \textit{position vector} $\mb{p}$ with its corresponding rotation matrix $\dot{\mb{R}} \mb{R}^\top$.

\vspace{5pt}
\begin{theorem}
Matrix $\dot{\mb{R}} \mb{R}^\top$ is \textit{skew symmetric}, {\ie},
\begin{equation}
(\dot{\mb{R}} \mb{R}^\top)^\top = - \dot{\mb{R}} \mb{R}^\top.
\end{equation}
\end{theorem}
\vspace{5pt}

\begin{proof}
We can prove it based on the what we learn from the previous Section~\ref{subsec:rotation_matrix_property} that matrix $\mb{R}$ is \textit{orthogonal}, {\ie}, $\mb{R}\mb{R}^\top = \mb{I}$. By taking the derivative of both sides of $\mb{R}$ is \textit{orthogonal}, {\ie}, $\mb{R}\mb{R}^\top = \mb{I}$ with time, we can get
\begin{equation}
\dot{\mb{R}} \mb{R}^\top + \mb{R} \dot{\mb{R}}^\top = \mb{0} \Rightarrow (\dot{\mb{R}} \mb{R}^\top)^\top = - \dot{\mb{R}} \mb{R}^\top,
\end{equation}
which concludes the proof that $\dot{\mb{R}} \mb{R}^\top$ is \textit{skew symmetric}.
\end{proof}

The readers may also wonder how can we represent $\dot{\mb{R}} \mb{R}^\top$? Actually, in Equation~\ref{equ:velocity_vector_representation}, we have provided one potential representation of matrix $\dot{\mb{R}} \mb{R}^\top$ with elements form the \textit{angular velocity vector} $\bs{\omega}$ already, {\ie},
\begin{equation}
\bs{\omega} \times \mb{p} = 
\underbrace{\begin{bmatrix}
0 & -\omega_z & \omega_y \\
\omega_z & 0 & -\omega_x \\
-\omega_y & \omega_x & 0 \\
\end{bmatrix}}_{\dot{\mb{R}} \mb{R}^\top}
\underbrace{\begin{bmatrix}
p_x\\
p_y\\
p_z\\
\end{bmatrix}}_{\mb{p}}.
\end{equation}

\begin{definition}
(Projection Operators): To simplify the notations, we can also introduce two operators $\land: \bs{\omega} \to \dot{\mb{R}} \mb{R}^\top$ and $\lor: \dot{\mb{R}} \mb{R}^\top \to \bs{\omega}$. 
\begin{itemize}
\item Given an angular velocity vector $\bs{\omega}$, we can represent the corresponding matrix $\dot{\mb{R}} \mb{R}^\top$ projected from it as 
\begin{equation}\label{equ:omega_hat}
\dot{\mb{R}} \mb{R}^\top = \bs{\omega}^{\land} \text{, or } \dot{\mb{R}} \mb{R}^\top = \widehat{\bs{\omega}}.
\end{equation}

\item Given a \textit{skew symmetric} matrix $\dot{\mb{R}} \mb{R}^\top$, we can also represent the corresponding angular velocity vector $\bs{\omega}$ constructed from it as 
\begin{equation}
\bs{\omega} = (\dot{\mb{R}} \mb{R}^\top)^{\lor}.
\end{equation}
\end{itemize}
\end{definition}

Based on the above analysis, we can also rewrite Equation~\ref{equ:velocity_rotation_matrix_relationship_equation} as follows:
\begin{equation}\label{equ:angular_velocity_transformation}
\bs{\omega} \times \mb{p} = \widehat{\bs{\omega}} \mb{p},
\end{equation}
which can calculate the vector cross-product with the regular matrix-vector multiplication.

%------------------------------------------------------------------------
%------------------------------------------------------------------------

\subsection{Rotation Matrix Exponential and Logarithm}

As robot arm rotates with angular velocity, it will change the arm's position with the rotation matrix. It seems both the angular velocity vector $\bs{\omega}$ and the rotation matrix $\mb{R}$ are describing the effects of the robot rotation motion. Then, given the angular velocity vector $\bs{\omega}$, can we directly calculate the corresponding rotation matrix $\mb{R}$? And given the rotation matrix $\mb{R}$, can we calculate the corresponding angular velocity vector $\bs{\omega}$?

To answer these two questions, at the end of this section, we will introduce two important operators defined on the rotation matrix $\mb{R}$, {\ie}, the \textit{rotation matrix exponential} and \textit{rotation matrix logarithm}.

%------------------------------------------------------------------------

\subsubsection{Rotation Matrix Exponential}

For Equation~\ref{equ:omega_hat}, if we multiply both sides with the rotation matrix $\mb{R}$, we can get
\begin{equation}\label{equ:exponential_representation}
\widehat{\bs{\omega}} \mb{R} = \dot{\mb{R}} \underbrace{\mb{R}^\top \mb{R}}_{= \mb{I}} = \dot{\mb{R}}.
\end{equation}

From the calculus course we learn at college, for a function $f(x)$, if $f(x)' = a f(x)$, then the function $f(x)$ can be represented as
\begin{equation}
f(x) = \exp^{a x} f(0).
\end{equation}

Similar representation can also be applied to represent the matrix $\mb{R}$ subject to Equation~\ref{equ:exponential_representation}, we can denote the matrix $\mb{R}$ as
\begin{equation}
\mb{R}(t) = \exp^{\widehat{\bs{\omega}} t} \mb{R}(0).
\end{equation}
In the above equation, we clearly specify the matrix $\mb{R}$ has variable $t$ and $\mb{R}(0)$ denotes the representation of matrix $\mb{R}$ at timestamp $t=0$. If we assume the initial condition $\mb{R}(0) = \mb{I}$, the above representation can be further simplified as
\begin{equation}\label{equ:rotation_matrix_exponential}
\mb{R}(t) = \exp^{\widehat{\bs{\omega}} t}.
\end{equation}

Meanwhile, let's come back to the knowledge we learn from the calculus course, according to the Taylor series, the exponential term $\exp^{a x}$ we use to represent the $f(x)$ function can actually be calculated as the sum of a sequence of polynomial terms:
\begin{equation}
\exp^{a x} = 1 + a x + \frac{(ax)^2}{2!} + \frac{ax)^3}{3!} + \cdots
\end{equation}

Therefore, we can further represent the matrix $\mb{R}(t)$ shown above as the sum of a sequence of matrix polynomial terms
\begin{equation}\label{equ:taylor_series_rotation_matrix}
\mb{R}(t) = \mb{I} + \widehat{\bs{\omega}} t + \frac{(\widehat{\bs{\omega}} t)^2}{2!} + \frac{(\widehat{\bs{\omega}} t)^3}{3!} + \cdots.
\end{equation}

\vspace{10pt}
\begin{theorem}
Given the angular velocity vector $\bs{\omega}$, it can be represented as $\bs{\omega} = \dot{q} \mb{e}$, where $\dot{q}$ is a scalar denotes the rotation velocity and $\mb{e}$ is a unit vector representing the angular velocity direction. Then we can have $\widehat{\bs{\omega}} = \dot{q} \widehat{\mb{e}}$. For the unit direction vector $\mb{e}$, subject to the $\land$ operator, we have 
\begin{align}
\widehat{\mb{e}}^{2n + 1} = (-1)^n \cdot \widehat{\mb{e}},\\
\widehat{\mb{e}}^{2n + 2} = (-1)^n \cdot \widehat{\mb{e}}^2.
\end{align}
\end{theorem}
We will leave the proof of the above theorem as an exercise for the readers, which can be done with the mathematical induction on $n$ starting from $n=0$. 
\vspace{10pt}

Based on the above theorem, we can further simplify the Equation~\ref{equ:taylor_series_rotation_matrix}, the high-order power of the matrix $\widehat{\bs{\omega}}$ in the equation can be reduced to the sum of $\widehat{\mb{e}}$ and $\widehat{\mb{e}}^2$:
\begin{equation}
\mb{R}(t) = \mb{I} + \left( (\dot{q} t) - \frac{(\dot{q} t)^3}{3!} + \frac{(\dot{q} t)^5}{5!} - \cdots \right) \widehat{\mb{e}} + \left( \frac{(\dot{q} t)^2}{2!} - \frac{(\dot{q} t)^4}{4!} + \frac{(\dot{q} t)^6}{6!} - \cdots \right) \widehat{\mb{e}}^2 .
\end{equation}

If the readers still remember the Taylor series of $\sin \theta$ and $\cos \theta$, you will observe that the above representation of matrix $\mb{R}(t)$ can be further simplified as follows:
\begin{equation}\label{equ:rotation_matrix_from_angular_velocity}
\mb{R}(t) = \mb{I} + \widehat{\mb{e}} \sin \theta + \widehat{\mb{e}}^2 (1 - \cos \theta) ,
\end{equation}
where $\theta = \dot{q} t$ is introduced to further simplify the representations. This equation will be frequently used in robot control and kinematics to be introduced in the following articles.

%------------------------------------------------------------------------

\subsubsection{Rotation Matrix Logarithm}

Based on the analysis in the above subsection, given the object angular velocity vector $\widehat{\bs{\omega}}$, we can directly calculate its corresponding rotation matrix $\mb{R}$ with Equation~\ref{equ:rotation_matrix_from_angular_velocity}. Meanwhile, when given a rotation matrix $\mb{R}$, can we also directly calculate its corresponding angular velocity vector or not? This is what we plan to introduce in this part.

According to Equation~\ref{equ:rotation_matrix_exponential}, the rotation matrix $\mb{R}$ can be represented as $\exp^{\widehat{\bs{\omega}}}$ (the time variable $t$ is discarded). By defining the logarithm operator on the rotation matrix, we can calculate the angular velocity vector $\bs{\omega}$ to be
\begin{equation}
\bs{\omega} = (\widehat{\bs{\omega}})^{\lor} = (\ln \mb{R})^{\lor}.
\end{equation}

Depending on the shape and contents of the rotation matrix $\mb{R} = \begin{bmatrix}
r_{11} & r_{12} & r_{13} \\
r_{21} & r_{22} & r_{23} \\
r_{31} & r_{32} & r_{33} \\
\end{bmatrix}$, the actual calculation can be represented as
\begin{equation}
\bs{\omega} = (\ln \mb{R})^{\lor} = \begin{cases}
(0, 0, 0)^\top & \text{if } \mb{R} = \mb{I}\\
\frac{\pi}{2} (r_{11}+1, r_{22}+1, r_{33}+1, ) & \text{if } \mb{R} \textit{ is diagonal}\\
\theta \frac{\mb{a}}{\left| \mb{a} \right|} & \text{otherwise}.
\end{cases}
\end{equation}
In the above representation, the scalar $\theta$ and vector $\mb{a}$ are defined based on the elements in matrix $\mb{R}$ as follows:
\begin{equation}
\mb{a} = (r_{32}-r_{23}, r_{13}-r_{31}, r_{21}-r_{12})^\top , \theta = \mathrm{atan2}( \left| \mb{a} \right|, r_{11}+r_{22}+r_{33}-1),
\end{equation}
where $\mathrm{atan2}(y, x)$ is the 2-argument arctangent function and it returns the angle in radians between the vector $(x, y)$ (composed with the input arguments) and the $x$ axis in the 2D Cartesian plane.

%------------------------------------------------------------------------
\section{Robot End Point Velocity}\label{sec:robot_velocity_torque}

In the previous Section~\ref{subsec:angular_velocity}, we have introduce the object rotation \textit{angular velocity} and \textit{linear velocity}. For an object rotating around the axis, we have illustrated their relationships and also discussed about the transformation on them by the rotation movement. 

In this section, we will further discuss about the \textit{angular velocity} and \textit{linear velocity} about the end point of a robot during the rotational motion.

\subsection{Single-Link Robot End Point Linear Velocity} 

%------------------------
\begin{figure}[t]
    \centering
    \includegraphics[width=0.9\textwidth]{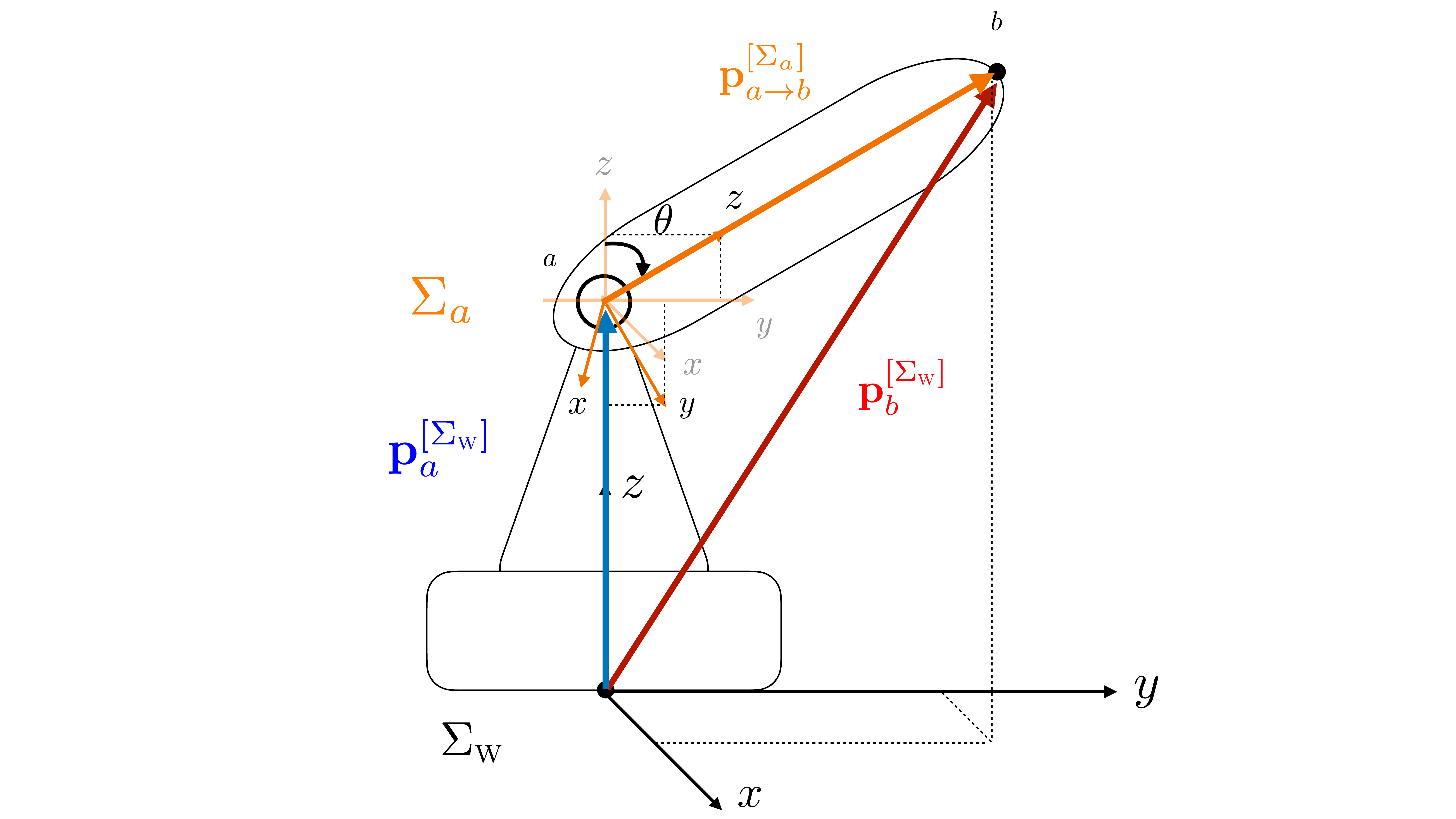}
    \caption{An Example of Single-Link Robot End Point Velocity.}
    \label{fig:local_configuration_example_2}
\end{figure}
%------------------------

We can borrow the example we use in the previous Section~\ref{sec:robot_representation} here again. As shown in Figure~\ref{fig:local_configuration_example_2}, given the robot arm with one movable link, we can represent its end point in the world coordinate system as
\begin{equation}\label{equ:single_arm_end_point_vector}
\mb{p}_b^{[\Sigma_{\textsc{w}}]} = \mb{p}_a^{[\Sigma_{\textsc{w}}]} + \mb{R}^{[\Sigma_{a} \to \Sigma_{\textsc{w}}]} \mb{p}_{a \to b}^{[\Sigma_{a}]}.
\end{equation}

In the above equation, vector $\mb{p}_{a \to b}^{[\Sigma_{a}]}$ pointing from joint $a$ to the end point $b$ within the local coordinate system $\Sigma_{a}$ is fixed and will not change with time. Meanwhile, the other vectors are all dynamic and will change as the robot moves. By calculating the derivatives of both sides of the above equation with time, we can obtain the \textit{linear velocity} of the end point $b$ as
\begin{align}
\mb{v}_b^{[\Sigma_{\textsc{w}}]} = \dot{\mb{p}}_b^{[\Sigma_{\textsc{w}}]}  & = \dot{\mb{p}}_a^{[\Sigma_{\textsc{w}}]} + \dot{\mb{R}}^{[\Sigma_{a} \to \Sigma_{\textsc{w}}]} \mb{p}_{a \to b}^{[\Sigma_{a}]} \\
& = \mb{v}_a^{[\Sigma_{\textsc{w}}]} + \left( \widehat{\bs{\omega}}_a^{[\Sigma_{\textsc{w}}]} \mb{R}^{[\Sigma_{a} \to \Sigma_{\textsc{w}}]} \right) \mb{p}_{a \to b}^{[\Sigma_{a}]}\\
& = \mb{v}_a^{[\Sigma_{\textsc{w}}]} +  \widehat{\bs{\omega}}_a^{[\Sigma_{\textsc{w}}]} \left( \mb{R}^{[\Sigma_{a} \to \Sigma_{\textsc{w}}]} \mb{p}_{a \to b}^{[\Sigma_{a}]} \right)\\
& =  \mb{v}_a^{[\Sigma_{\textsc{w}}]} + {\bs{\omega}}_a^{[\Sigma_{\textsc{w}}]} \times  \mb{p}_{a \to b}^{[\Sigma_{\textsc{w}}]} \\
& =  \mb{v}_a^{[\Sigma_{\textsc{w}}]} + {\bs{\omega}}_a^{[\Sigma_{\textsc{w}}]} \times \left( {\mb{p}}_b^{[\Sigma_{\textsc{w}}]} - {\mb{p}}_a^{[\Sigma_{\textsc{w}}]} \right).
\end{align}
In the above equation, we use many results obtained from the previous sections: (1) according to Equation~\ref{equ:exponential_representation}, $\dot{\mb{R}}=\widehat{\bs{\omega}} \mb{R}$, (2) according to Equation~\ref{equ:angular_velocity_transformation}, $\widehat{\bs{\omega}} \mb{p}=\bs{\omega} \times \mb{p}$, and (3) according to the above Equation~\ref{equ:single_arm_end_point_vector}, $\mb{R}^{[\Sigma_{a} \to \Sigma_{\textsc{w}}]} \mb{p}_{a \to b}^{[\Sigma_{a}]}= \mb{p}_{a \to b}^{[\Sigma_{\textsc{w}}]} = \mb{p}_b^{[\Sigma_{\textsc{w}}]}-\mb{p}_a^{[\Sigma_{\textsc{w}}]}$.

Since there exist one single link that rotates, we can easily get that the \textit{angular velocity} of the end point will be equal to the velocity of the joint $a$, {\ie},
\begin{equation}
\bs{\omega}_b^{\Sigma_{\textsc{w}}} = \bs{\omega}_a^{\Sigma_{\textsc{w}}}.
\end{equation}

\subsection{Multi-Link Robot End Point Velocity} 

%------------------------
\begin{figure}[t]
    \centering
    \includegraphics[width=0.9\textwidth]{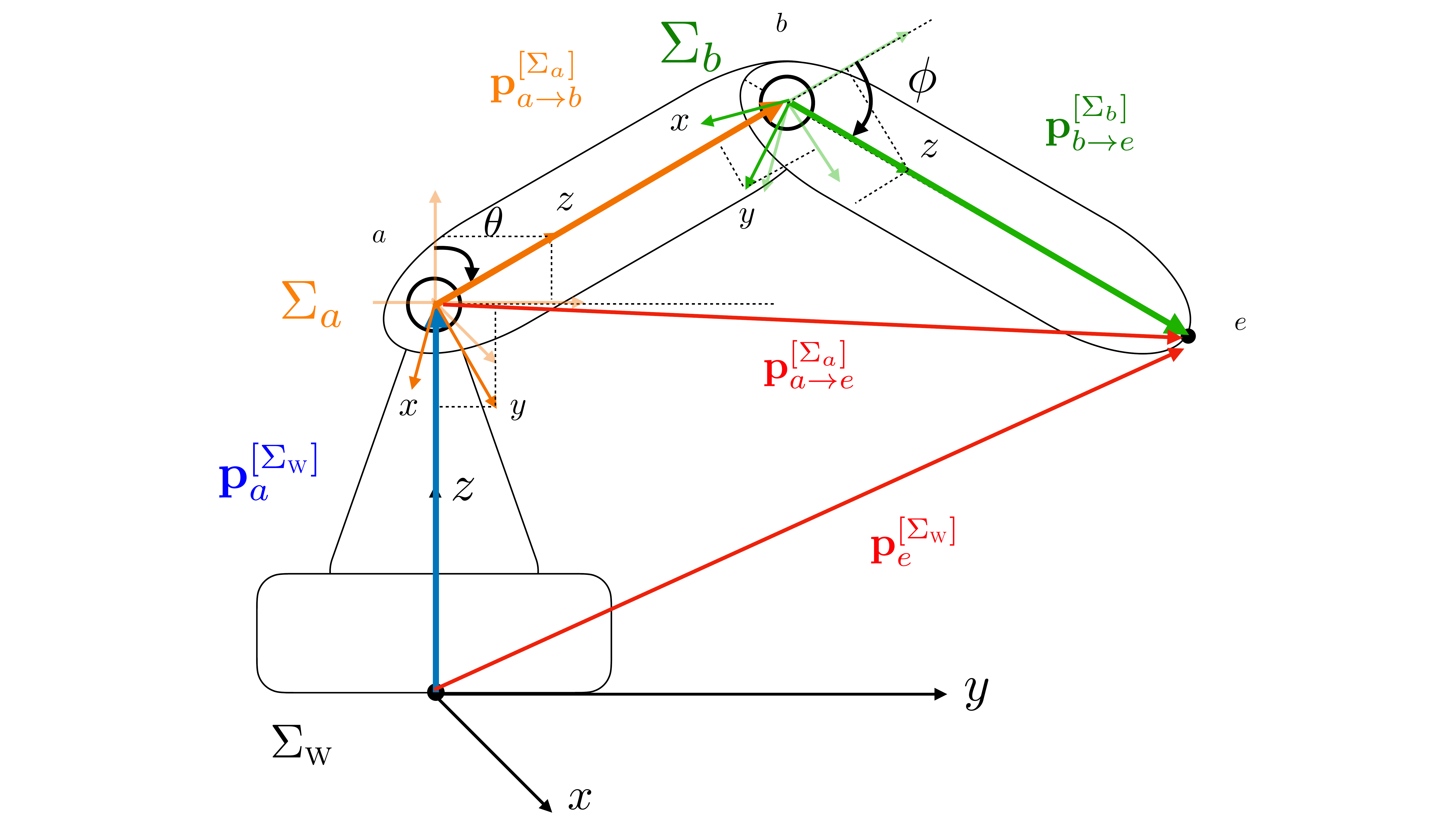}
    \caption{An Example of Multi-Link Robot End Point Velocity.}
    \label{fig:multi_link_robot_end_point_velocity}
\end{figure}
%------------------------

For the robot with multiple (more than one) moveable links, the end point's velocity calculation will be more difficulty and challenging. We take the multi-link robot arm used before in this part to illustrate the calculation process for readers.

\subsubsection{Multi-Link Robot Homogeneous Transition Matrix}

As introduced in Section~\ref{subsec:multi_link_robot_transformation}, for the robot arm involving multiple link, the \textit{homogenous transformation matrices} between the local and world coordinate systems can be represented as follows:
\begin{equation}
\mb{T}^{[\Sigma_b\to\Sigma_a]} = \begin{bmatrix}
\mb{R}^{[\Sigma_b\to\Sigma_a]} & \mb{p}_{a \to b}^{[\Sigma_a]}\\
\mb{0} & 1
\end{bmatrix} \text{, and } \mb{T}^{[\Sigma_{a} \to \Sigma_{\textsc{w}}]} = \begin{bmatrix}
\mb{R}^{[\Sigma_{a} \to \Sigma_{\textsc{w}}]} & \mb{p}_a^{[\Sigma_{\textsc{w}}]}\\
\mb{0} & 1
\end{bmatrix}
\end{equation}

Based on them, we can represent the \textit{homogenous transformation matrices} from the second arm link's local coordinate $\Sigma_b$ to the world coordinate $\Sigma_{\textsc{w}}$ as follows:
\begin{align}
\mb{T}^{[\Sigma_{b} \to \Sigma_{\textsc{w}}]} &= \mb{T}^{[\Sigma_{a} \to \Sigma_{\textsc{w}}]} \mb{T}^{[\Sigma_b\to\Sigma_a]} \\
&= \begin{bmatrix}
\mb{R}^{[\Sigma_{a} \to \Sigma_{\textsc{w}}]} & \mb{p}_a^{[\Sigma_{\textsc{w}}]}\\
\mb{0} & 1
\end{bmatrix} 
\begin{bmatrix}
\mb{R}^{[\Sigma_b\to\Sigma_a]} & \mb{p}_{a \to b}^{[\Sigma_a]}\\
\mb{0} & 1
\end{bmatrix} \\
&= \begin{bmatrix}
\left(\mb{R}^{[\Sigma_{a} \to \Sigma_{\textsc{w}}]} \mb{R}^{[\Sigma_b\to\Sigma_a]}\right) & \left(\mb{R}^{[\Sigma_{a} \to \Sigma_{\textsc{w}}]} \mb{p}_{a \to b}^{[\Sigma_a]} + \mb{p}_a^{[\Sigma_{\textsc{w}}]} \right)\\
\mb{0} & 1
\end{bmatrix}.
\end{align}

According to the physical meanings of elements in the calculated \textit{homogeneous transformation matrix} $\mb{T}^{[\Sigma_{b} \to \Sigma_{\textsc{w}}]}$, we know that the \textit{rotation matrix} from local coordinate $\Sigma_b$ at the second arm link to the world coordinate $\Sigma_{\textsc{w}}$ and the arm end point position vector in the world coordinate $\Sigma_{\textsc{w}}$ can be represented as
\begin{align}
&\mb{p}_b^{[\Sigma_{\textsc{w}}]} = \mb{R}^{[\Sigma_{a} \to \Sigma_{\textsc{w}}]} \mb{p}_{a \to b}^{[\Sigma_a]} + \mb{p}_a^{[\Sigma_{\textsc{w}}]}, \label{equ:multi_link_robot_position_vector}\\
&\mb{R}^{[\Sigma_{b} \to \Sigma_{\textsc{w}}]} = \mb{R}^{[\Sigma_{a} \to \Sigma_{\textsc{w}}]} \mb{R}^{[\Sigma_b\to\Sigma_a]},\label{equ:multi_link_robot_rotation_matrix}
\end{align}
which will help calculate the \textit{linear velocity} and \textit{angular velocity} vectors of the end point in the world coordinate $\Sigma_{\textsc{w}}$.

\subsubsection{Multi-Link Robot Linear Velocity}

From Equation~\ref{}, by projecting the second arm link vector $\mb{p}_{b \to e}^{[\Sigma_b]}$ and adding to the position vector $\mb{p}_b^{[\Sigma_{\textsc{w}}]}$, we can calculate the end point \textit{linear velocity} as
\begin{align}
\mb{p}_e^{[\Sigma_{\textsc{w}}]} &= \mb{p}_b^{[\Sigma_{\textsc{w}}]} + \mb{R}^{[\Sigma_{b} \to \Sigma_{\textsc{w}}]} \mb{p}_{b \to e}^{[\Sigma_b]} \\
&= \mb{p}_a^{[\Sigma_{\textsc{w}}]} + \mb{R}^{[\Sigma_{a} \to \Sigma_{\textsc{w}}]} \mb{p}_{a \to b}^{[\Sigma_a]} + \mb{R}^{[\Sigma_{a} \to \Sigma_{\textsc{w}}]} \mb{R}^{[\Sigma_b\to\Sigma_a]} \mb{p}_{b \to e}^{[\Sigma_b]}.
\end{align}

By taking the derivatives for both sides of Equation~\ref{equ:multi_link_robot_position_vector}, we can obtain the end point \textit{linear velocity} to be
\begin{align}
\small
\mb{v}_e^{[\Sigma_{\textsc{w}}]} =& \dot{\mb{p}}_e^{[\Sigma_{\textsc{w}}]} \\
=& \dot{\mb{p}}_a^{[\Sigma_{\textsc{w}}]} + \dot{\mb{R}}^{[\Sigma_{a} \to \Sigma_{\textsc{w}}]} \mb{p}_{a \to b}^{[\Sigma_a]} + \dot{\mb{R}}^{[\Sigma_{a} \to \Sigma_{\textsc{w}}]} \mb{R}^{[\Sigma_b\to\Sigma_a]} \mb{p}_{b \to e}^{[\Sigma_b]} + \mb{R}^{[\Sigma_{a} \to \Sigma_{\textsc{w}}]} \dot{\mb{R}}^{[\Sigma_b\to\Sigma_a]} \mb{p}_{b \to e}^{[\Sigma_b]} \\
=& \mb{v}_a^{[\Sigma_{\textsc{w}}]} + \widehat{\bs{\omega}}_a^{[\Sigma_{\textsc{w}}]} {\mb{R}}^{[\Sigma_{a} \to \Sigma_{\textsc{w}}]} \mb{p}_{a \to b}^{[\Sigma_a]} + \widehat{\bs{\omega}}_a^{[\Sigma_{\textsc{w}}]} {\mb{R}}^{[\Sigma_{a} \to \Sigma_{\textsc{w}}]} \mb{R}^{[\Sigma_b\to\Sigma_a]} \mb{p}_{b \to e}^{[\Sigma_b]} \\
&+ \mb{R}^{[\Sigma_{a} \to \Sigma_{\textsc{w}}]} \widehat{\bs{\omega}}_b^{[\Sigma_a]} {\mb{R}}^{[\Sigma_b\to\Sigma_a]} \mb{p}_{b \to e}^{[\Sigma_b]} \\
=& \mb{v}_a^{[\Sigma_{\textsc{w}}]} + \widehat{\bs{\omega}}_a^{[\Sigma_{\textsc{w}}]} \left( {\mb{R}}^{[\Sigma_{a} \to \Sigma_{\textsc{w}}]} \mb{p}_{a \to b}^{[\Sigma_a]} \right) + \widehat{\bs{\omega}}_a^{[\Sigma_{\textsc{w}}]} \left( {\mb{R}}^{[\Sigma_{a} \to \Sigma_{\textsc{w}}]} \mb{R}^{[\Sigma_b\to\Sigma_a]} \mb{p}_{b \to e}^{[\Sigma_b]} \right) \\
&+ \mb{R}^{[\Sigma_{a} \to \Sigma_{\textsc{w}}]} \widehat{\bs{\omega}}_b^{[\Sigma_a]} \left( {\mb{R}}^{[\Sigma_b\to\Sigma_a]} \mb{p}_{b \to e}^{[\Sigma_b]} \right) \\
=& \mb{v}_a^{[\Sigma_{\textsc{w}}]} + {\bs{\omega}}_a^{[\Sigma_{\textsc{w}}]} \times \mb{p}_{a \to b}^{[\Sigma_{\textsc{w}}]} + {\bs{\omega}}_a^{[\Sigma_{\textsc{w}}]} \times \mb{p}_{b \to e}^{[\Sigma_{\textsc{w}}]} + \mb{R}^{[\Sigma_{a} \to \Sigma_{\textsc{w}}]} \left( {\bs{\omega}}_b^{[\Sigma_a]} \times \mb{p}_{b \to e}^{[\Sigma_a]} \right) \\
=& \mb{v}_a^{[\Sigma_{\textsc{w}}]} + {\bs{\omega}}_a^{[\Sigma_{\textsc{w}}]} \times \left(\mb{p}_{a \to b}^{[\Sigma_{\textsc{w}}]} + \mb{p}_{b \to e}^{[\Sigma_{\textsc{w}}]} \right) + \left( \mb{R}^{[\Sigma_{a} \to \Sigma_{\textsc{w}}]} {\bs{\omega}}_b^{[\Sigma_a]} \right) \times \left(\mb{R}^{[\Sigma_{a} \to \Sigma_{\textsc{w}}]} \mb{p}_{b \to e}^{[\Sigma_a]} \right) \\
=& \mb{v}_a^{[\Sigma_{\textsc{w}}]} + {\bs{\omega}}_a^{[\Sigma_{\textsc{w}}]} \times \mb{p}_{a \to e}^{[\Sigma_{\textsc{w}}]} + {\bs{\omega}}_b^{[\Sigma_{\textsc{w}}]} \times \mb{p}_{b \to e}^{[\Sigma_{\textsc{w}}]} \\
=& \mb{v}_a^{[\Sigma_{\textsc{w}}]} + {\bs{\omega}}_a^{[\Sigma_{\textsc{w}}]} \times \left( \mb{p}_{e}^{[\Sigma_{\textsc{w}}]} -\mb{p}_{a}^{[\Sigma_{\textsc{w}}]} \right) + {\bs{\omega}}_b^{[\Sigma_{\textsc{w}}]} \times \left( \mb{p}_{e}^{[\Sigma_{\textsc{w}}]} -\mb{p}_{b}^{[\Sigma_{\textsc{w}}]} \right).
\end{align}

\subsubsection{Multi-Link Robot Angular Velocity}

Meanwhile, for the \textit{angular velocity} of the end point, since the end point $e$ is attached to the arm link in the coordinate system $\Sigma_b$ with origin at joint $b$, we can calculate it based on the rotation matrix $\mb{R}^{[\Sigma_{b} \to \Sigma_{\textsc{w}}]}$ introduced in Equation~\ref{equ:multi_link_robot_rotation_matrix}. As introduced before in Equation~\ref{equ:omega_hat}, we know that
\begin{align}
\widehat{\bs{\omega}}_e^{[\Sigma_{\textsc{w}}]} 
=& \dot{\mb{R}}^{[\Sigma_{b} \to \Sigma_{\textsc{w}}]} \left(\mb{R}^{[\Sigma_{b} \to \Sigma_{\textsc{w}}]}\right)^\top \\
=& \frac{d}{dt} \left( \mb{R}^{[\Sigma_{a} \to \Sigma_{\textsc{w}}]} \mb{R}^{[\Sigma_b\to\Sigma_a]} \right) \left(\mb{R}^{[\Sigma_{b} \to \Sigma_{\textsc{w}}]}\right)^\top \\
=& \left( \dot{\mb{R}}^{[\Sigma_{a} \to \Sigma_{\textsc{w}}]} \mb{R}^{[\Sigma_b\to\Sigma_a]} + \mb{R}^{[\Sigma_{a} \to \Sigma_{\textsc{w}}]} \dot{\mb{R}}^{[\Sigma_b\to\Sigma_a]} \right) \left(\mb{R}^{[\Sigma_{b} \to \Sigma_{\textsc{w}}]}\right)^\top \\
=& \left( \widehat{\bs{\omega}}_a^{[\Sigma_{\textsc{w}}]} {\mb{R}}^{[\Sigma_{a} \to \Sigma_{\textsc{w}}]} \mb{R}^{[\Sigma_b\to\Sigma_a]} + {\mb{R}}^{[\Sigma_{a} \to \Sigma_{\textsc{w}}]} \widehat{\bs{\omega}}_b^{[\Sigma_a]} \mb{R}^{[\Sigma_b\to\Sigma_a]}  \right) \left(\mb{R}^{[\Sigma_{b} \to \Sigma_{\textsc{w}}]}\right)^\top \\
=& \widehat{\bs{\omega}}_a^{[\Sigma_{\textsc{w}}]} \left( {\mb{R}}^{[\Sigma_{a} \to \Sigma_{\textsc{w}}]} \mb{R}^{[\Sigma_b\to\Sigma_a]} \right) \left(\mb{R}^{[\Sigma_{b} \to \Sigma_{\textsc{w}}]}\right)^\top\\
& + {\mb{R}}^{[\Sigma_{a} \to \Sigma_{\textsc{w}}]} \widehat{\bs{\omega}}_b^{[\Sigma_a]} \mb{R}^{[\Sigma_b\to\Sigma_a]} \left(\mb{R}^{[\Sigma_{a} \to \Sigma_{\textsc{w}}]} \mb{R}^{[\Sigma_b\to\Sigma_a]} \right)^\top\\
=& \widehat{\bs{\omega}}_a^{[\Sigma_{\textsc{w}}]} +  {\mb{R}}^{[\Sigma_{a} \to \Sigma_{\textsc{w}}]} \widehat{\bs{\omega}}_b^{[\Sigma_a]} \mb{R}^{[\Sigma_b\to\Sigma_a]}  \left(\mb{R}^{[\Sigma_b\to\Sigma_a]} \right)^\top \left(\mb{R}^{[\Sigma_{a} \to \Sigma_{\textsc{w}}]}\right)^\top\\
=& \widehat{\bs{\omega}}_a^{[\Sigma_{\textsc{w}}]} +  {\mb{R}}^{[\Sigma_{a} \to \Sigma_{\textsc{w}}]} \widehat{\bs{\omega}}_b^{[\Sigma_a]} \left(\mb{R}^{[\Sigma_{a} \to \Sigma_{\textsc{w}}]}\right)^\top \label{equ:step1}\\
=& \widehat{\bs{\omega}}_a^{[\Sigma_{\textsc{w}}]} +  \left( {\mb{R}}^{[\Sigma_{a} \to \Sigma_{\textsc{w}}]} {\bs{\omega}}_b^{[\Sigma_a]} \right)^{\land} \label{equ:step2}\\
=& \widehat{\bs{\omega}}_a^{[\Sigma_{\textsc{w}}]} + \widehat{\bs{\omega}}_b^{[\Sigma_{\textsc{w}}]}.
\end{align}

By applying the $\lor$ operator to both sides of the above equation, we can represent the \textit{angular velocity} of the end point as
\begin{equation}
{\bs{\omega}}_e^{[\Sigma_{\textsc{w}}]} = {\bs{\omega}}_a^{[\Sigma_{\textsc{w}}]} + {\bs{\omega}}_b^{[\Sigma_{\textsc{w}}]}.
\end{equation}

To derive the \textit{angular velocity} and \textit{linear velocity} of the end point, we use several equations to replace terms with their equivalent representations, like 
\begin{equation}
\mb{R}^{[\Sigma_{a} \to \Sigma_{\textsc{w}}]} \left( {\bs{\omega}}_b^{[\Sigma_a]} \times \mb{p}_{b \to e}^{[\Sigma_a]} \right) = \left( \mb{R}^{[\Sigma_{a} \to \Sigma_{\textsc{w}}]} {\bs{\omega}}_b^{[\Sigma_a]} \right) \times \left(\mb{R}^{[\Sigma_{a} \to \Sigma_{\textsc{w}}]} \mb{p}_{b \to e}^{[\Sigma_a]} \right)
\end{equation} 
and
\begin{equation}
{\mb{R}}^{[\Sigma_{a} \to \Sigma_{\textsc{w}}]} \widehat{\bs{\omega}}_b^{[\Sigma_a]} \left(\mb{R}^{[\Sigma_{a} \to \Sigma_{\textsc{w}}]}\right)^\top = \left( {\mb{R}}^{[\Sigma_{a} \to \Sigma_{\textsc{w}}]} {\bs{\omega}}_b^{[\Sigma_a]} \right)^{\land}. 
\end{equation} 
We will leave the proof of the above two equations as an exercise for the readers.

\subsection{A General Representation}

Based on the above analysis, given the robot arm with multi-links, whose rotation joints are denoted as $a$, $b$, $c$, $\cdots$, we can represent the end point's \textit{linear velocity} and \textit{angular velocity} of the robot arm as follows:
\begin{align}
\textbf{Angular Velocity: } {\bs{\omega}}_e^{[\Sigma_{\textsc{w}}]} = &\   {\bs{\omega}}_a^{[\Sigma_{\textsc{w}}]} + {\bs{\omega}}_b^{[\Sigma_{\textsc{w}}]} + {\bs{\omega}}_c^{[\Sigma_{\textsc{w}}]} + \cdots,\\
\textbf{Linear Velocity: }\mb{v}_e^{[\Sigma_{\textsc{w}}]} = &\ \mb{v}_a^{[\Sigma_{\textsc{w}}]} + {\bs{\omega}}_a^{[\Sigma_{\textsc{w}}]} \times \left( \mb{p}_{e}^{[\Sigma_{\textsc{w}}]} -\mb{p}_{a}^{[\Sigma_{\textsc{w}}]} \right) \\
&+ {\bs{\omega}}_b^{[\Sigma_{\textsc{w}}]} \times \left( \mb{p}_{e}^{[\Sigma_{\textsc{w}}]} -\mb{p}_{b}^{[\Sigma_{\textsc{w}}]} \right) \\
&+{\bs{\omega}}_c^{[\Sigma_{\textsc{w}}]} \times \left( \mb{p}_{e}^{[\Sigma_{\textsc{w}}]} -\mb{p}_{c}^{[\Sigma_{\textsc{w}}]} \right) + \cdots.
\end{align}

%------------------------------------------------------------------------
\section{What's Next?}

By now, we have introduced the robot representations, robot rotation, and discuss about the robot end point velocity after the rotation. In the next article, we will further talk about several advanced topics about robot motion and control, including \textit{forward kinematics}, \textit{inverse kinematics}, \textit{trajectory generation} and \textit{motion planning}.
%------------------------------------------------------------------------

\newpage

\vskip 0.2in
\bibliographystyle{plain}
\bibliography{reference}

\begin{thebibliography}{1}

\bibitem{10.5555/3100040}
Shuuji Kajita, Hirohisa Hirukawa, Kensuke Harada, and Kazuhito Yokoi.
\newblock {\em Introduction to Humanoid Robotics}.
\newblock Springer Publishing Company, Incorporated, 1st edition, 2016.

\bibitem{10.5555/3165183}
Kevin~M. Lynch and Frank~C. Park.
\newblock {\em Modern Robotics: Mechanics, Planning, and Control}.
\newblock Cambridge University Press, USA, 1st edition, 2017.

\bibitem{un_ageing}
United Nations.
\newblock Global issues: Ageing.

\end{thebibliography}

\end{document}